# Modified EDAS Method Based on Cumulative Prospect Theory for Multiple Attributes Group Decision Making with Interval-valued Intuitionistic Fuzzy Information

Jing Wang[1], Qiang Cai[2], Guiwu Wei[1,2*], Ningna Liao[2]

[1]School of Mathematical Sciences, Sichuan Normal University, Chengdu, 610101, P.R. China
[2]School of Business, Sichuan Normal University, Chengdu, 610101, P.R. China

*Correspondence: weiguiwu1973@sicnu.edu.cn

**Abstract:** The Interval-valued intuitionistic fuzzy sets (IVIFSs) based on the intuitionistic fuzzy sets combines the classical decision method is in its research and application is attracting attention. After comparative analysis, there are multiple classical methods with IVIFSs information have been applied into many practical issues. In this paper, we extended the classical EDAS method based on cumulative prospect theory (CPT) considering the decision makers (DMs)' psychological factor under IVIFSs. Taking the fuzzy and uncertain character of the IVIFSs and the psychological preference into consideration, the original EDAS method based on the CPT under IVIFSs (IVIF-CPT-MABAC) method is built for MAGDM issues. Meanwhile, information entropy method is used to evaluate the attribute weight. Finally, a numerical example for project selection of green technology venture capital has been given and some comparisons is used to illustrate advantages of IVIF-CPT-MABAC method and some comparison analysis and sensitivity analysis are applied to prove this new method's effectiveness and stability.

## 1. Introduction

In most practical decisions, the decision maker's evaluation of an index can't be simply expressed by real numbers [1-4]. Therefore, Zadeh [5] introduced fuzzy sets (FSs) and then this theory and its extension have been applied to many fields [6-9]. The extension of the fuzzy set (FS) such as the intuitionistic fuzzy set (IFS)[10] is the classical fuzzy set, and constrained by the fact that the IFS is in the range of real number, the interval-valued intuitionistic sets (IVIFSs) extend the IFS with the interval numbers to express different condition in decision-making. Atanassov and Gargov [11] proposed the IVIFSs, and the corresponding operators[12], and with the basic research set up, the corresponding innovation of IVIFSs has been proposed. Grzegorzewski [13] proposed a new distance assessment based on Hausdorff metric which is the well-known Hamming distance. Based on some operational laws of IVIFSs, Meanwhile, TOPSIS method[14-18], grey relational analysis (GRA) method[19-22], VIKOR method[23-29], TODIM method[30-35], MABAC method[36-39], and other methods have been applied in the IVIFSs[40-45].

EDAS method [46, 47]assesses the alternatives by the positive (negative) distance from average (PDA and NDA). In other words, the higher value of the PDA or the lower value of the NDA means a more optimal alternative. Compared with other methods, the EDAS method has much easier formulas so that it is more effective than others. Thus, EDAS method has been applied into many fuzzy sets. For instance, IFSs[48], interval-valued fuzzy soft sets[49], hesitant fuzzy linguistic sets [50], interval-valued neutrosophic sets[51], and so on. Thus, it can be seen that EDAS method has good adaptability in the decision-making fields. Though Li and Wang [52] extended the EDAS method in IVIFSs, this model can't take the risk reference of DM as consideration. Due to the fact that most methods are based on the utility theory which consider that DMs are rational, the combination of CPT and classical EDAS method are now applied to many fuzzy sets, such as probabilistic hesitant fuzzy sets[53] and picture fuzzy sets[54].

Thus, considering the simple calculation of EDAS method, the CPT's characteristic and the IVIFSs which may contain more information, the IVIF-CPT-EDAS method is constructed in this paper. Therefore, it is necessary to integrate the CPT which takes the DMs' risk preference into consideration and classical MAGDM methods. The aim of this paper is to extend EDAS method based on CPT for MAGDM under IVIFSs and apply it into project selection of green technology venture capital. The contributions of this paper are shown as follows: (1) The integration of CPT and EDAS method in IVIFSs which not only considers relatively simple and reasonable classical method, but also considers the psychological state of DMs which is more realistic has been constructed. (2) The information entropy method is extended into IVIFSs to calculate the weight of attributes. (3) the IVIF-CPT- EDAS method is proposed for MAGDM issues. (4) Finally, a numerical example for project selection of green technology venture capital has been given and some comparisons is used to illustrate advantages of IVIF-EDAS-MABAC method and some comparison analysis and sensitivity analysis are applied to prove this new method's effectiveness and stability. In order to do so, the overall structure of the article is as follows: the basic knowledges of IVIFSs are introduced in Section 2, and the IVIF-CPT-EDAS method is constructed in Section 3. Section 4 gave a numerical study for project selection of green technology venture capital, and the corresponding parameters' sensitivity, and in the Section 5, comparison analysis is then to prove its effectively and stability. In Section 6, the conclusion is then made to summarize this paper.

## 2. Preliminary Knowledges

Since the fuzzy sets (FSs) introduced by Zadeh[5] in 1965, the FSs related theory has achieved rich results through the joint efforts of many scholars. Especially, the intuitionistic fuzzy sets (IFSs) proposed by Atanassov[10] in 1986 and the interval-valued intuitionistic fuzzy sets (IVIFSs) proposed by Atanassov and Gargov [11] in 1989 to extend the original theory. Therefore, in this section, the related knowledges of FSs and its extension is then introduced to provide the theoretical basis for MADM problems.

### 2.1 IVIFSs

**Definition 1[5].** Set $T$ as a finitely non-empty set, and the FSs on $T$ is described as follows:

$$FS = \left\{ \left\langle t, \eta_{IFS}\left(t\right) \right\rangle \middle| t \in T \right\} \tag{2.1}$$

where $\eta_{IFS}\left(t\right)$ denotes the membership of FSs, and $\eta_{IFS} : T \to \left[0,1\right]$ is a single real value.

**Definition 2[10].** Set $T$ as a finitely non-empty set, and the IFSs on $T$ is described as follows:

$$IFS = \left\{ \left\langle t, \eta_{IFS}\left(t\right), v_{IFS}\left(t\right) \right\rangle \middle| t \in T \right\} \tag{2.2}$$

where $\eta_{IFS}\left(t\right)$ and $v_{IFS}\left(t\right)$ represents the membership and non-membership of IFSs and is denoted as follows:

$$\eta_{IFS}\left(t\right) : T \to \left[0,1\right], t \in T \to \eta_{IFS}\left(t\right) \in \left[0,1\right] \tag{2.3}$$

$$v_{IFS}\left(t\right) : T \to \left[0,1\right], t \in T \to v_{IFS}\left(t\right) \in \left[0,1\right] \tag{2.4}$$

$$0 \le \eta_{IFS}\left(t\right) + v_{IFS}\left(t\right) \le 1 \tag{2.5}$$

Besides, the hesitation degree of IFSs is denoted as follows:

$$\pi_{IFS}\left(t\right) = 1 - \eta_{IFS}\left(t\right) - v_{IFS}\left(t\right) \tag{2.6}$$

where $\forall t \in T$, $0 \le \pi_{IFS}\left(t\right) \le 1$.

**Definition 3[11].** Set $T$ as a finitely non-empty set, and the IVIFSs on $T$ is described as follows:

$$IVIFS = \left\{ \left\langle t, \tilde{\eta}_{IVIFS}(t), \tilde{\nu}_{IVIFS}(t) \right\rangle \middle| t \in T \right\} \tag{2.7}$$

where $\tilde{\eta}_{IVIFS}(t)$ and $\tilde{\nu}_{IVIFS}(t)$ represents the interval of membership and non-membership which is shown as follows:

$$\tilde{\eta}_{IVIFS}(t) : t \in T \to \tilde{\eta}_{IVIFS}(t) = \left[ LM(t), RM(t) \right] \subseteq [0,1] \tag{2.8}$$

$$\tilde{\nu}_{IVIFS}(t) : t \in T \to \tilde{\nu}_{IVIFS}(t) = \left[ LN(t), RN(t) \right] \subseteq [0,1] \tag{2.9}$$

$$0 \le RM(t) + RN(t) \le 1 \tag{2.10}$$

Similarly, the interval of hesitation of IVIFSs is described as follows:

$$\tilde{\pi}_{IVIFS}(t) = \left[ 1 - RM(t) + RN(t), 1 - LM(t) - LN(t) \right] \tag{2.11}$$

where $\forall t \in T$, $0 \le \tilde{\pi}_{IVIFS}(t) \le 1$.

Similarly, the Interval-valued intuitionistic fuzzy number (IVIFN)[55] of IVIFSs in Eq. (2.6) is shown in Eq. (2.12) .

$$IVIFN = \left( \tilde{\eta}_{IVIFS}(t), \tilde{\nu}_{IVIFS}(t) \right) = \left( \left[ LM(t), RM(t) \right], \left[ LN(t), RN(t) \right] \right) \tag{2.12}$$

The Eq. (2.12) can be abbreviated in Eq. (2.3):

$$IVIFN = \left( \left[ LM, RM \right], \left[ LN, RN \right] \right) \tag{2.13}$$

Specially, when $LM = RM, LN = RN$ , IVIFSs can be degenerated into IFSs. The maximum IVIFN is $IVIFN_{\max} = \left( [1,1], [0,0] \right)$, and the minimum IVIFN is $IVIFN_{\min} = \left( [0,0], [1,1] \right)$.

**Definition 4[56].** Suppose any three IVIFNs $IVIFN_i = \left( \left[ LM_i, RM_i \right], \left[ LN_i, RN_i \right] \right)$, $(i = 1,2,3)$, and the calculation rules of IVIFNs are defined as follows:

(1) $\left( IVIFN_1 \right)^c = \left( \left[ LN_1, RN_1 \right], \left[ LM_1, RM_1 \right] \right)$

(2) $IVIFN_1 \oplus IVIFN_2 = \begin{pmatrix} \left[ LM_1 + LM_2 - LM_1 LM_2, RM_1 + RM_2 - RM_1 RM_2 \right], \\ \left[ LN_1 LN_2, RN_1 RN_2 \right] \end{pmatrix}$

(3) $IVIFN_1 \otimes IVIFN_2 = \begin{pmatrix} \left[ LM_1 LM_2, RM_1 RM_2 \right], \\ \left[ LN_1 + LN_2 - LN_1 LN_2, RN_1 + RN_2 - RN_1 RN_2 \right] \end{pmatrix}$

(4) $$IVIFN_1 \cup IVIFN_2 = \begin{pmatrix} \left[\max\left\{LM_1, LM_2\right\}, \max\left\{RM_1, RM_2\right\}\right] \\ ,\left[\min\left\{LN_1, LN_2\right\}, \min\left\{RN_1, RN_2\right\}\right] \end{pmatrix}$$

(5) $$IVIFN_1 \cap IVIFN_2 = \begin{pmatrix} \left[\min\left\{LM_1, LM_2\right\}, \min\left\{RM_1, RM_2\right\}\right] \\ ,\left[\max\left\{LN_1, LN_2\right\}, \max\left\{RN_1, RN_2\right\}\right] \end{pmatrix}$$

(6) $$\kappa IVIFN_1 = \left(\left[1-\left(1-LM_1\right)^{\kappa}, 1-\left(1-RM_1\right)^{\kappa}\right], \left[LN_1^{\kappa}, RN_1^{\kappa}\right]\right), \kappa > 0$$

(7) $$IVIFN_1^{\kappa} = \left(\left[LM_1^{\kappa}, RM_1^{\kappa}\right], \left[1-\left(1-LN_1\right)^{\kappa}, 1-\left(1-RN_1\right)^{\kappa}\right]\right), \kappa > 0$$

Besides, the above algorithm also satisfies the following operation laws[57]:

i. Commutative law

a) $IVIFN_1 \oplus IVIFN_2 = IVIFN_2 \oplus IVIFN_1$

b) $IVIFN_1 \otimes IVIFN_2 = IVIFN_2 \otimes IVIFN_1$

ii. Associative law

a) $\left(IVIFN_1 \oplus IVIFN_2\right) \oplus IVIFN_3 = IVIFN_1 \oplus \left(IVIFN_2 \oplus IVIFN_3\right)$

b) $\left(IVIFN_1 \otimes IVIFN_2\right) \otimes IVIFN_3 = IVIFN_1 \otimes \left(IVIFN_2 \otimes IVIFN_3\right)$

iii. Distributive law

a) $\kappa\left(IVIFN_1 \oplus IVIFN_2\right) = \kappa IVIFN_1 \oplus \kappa IVIFN_2$

b) $\kappa_1 IVIFN_1 \oplus \kappa_2 IVIFN_1 = \left(\kappa_1 + \kappa_2\right) IVIFN_1$

iv. Exponential operation law

a) $\left(IVIFN_1 \otimes IVIFN_2\right)^{\kappa} = IVIFN_1^{\kappa} \otimes IVIFN_2^{\kappa}$

b) $IIFN_1^{\kappa_1} \otimes IIFN_1^{\kappa_2} = IIFN_1^{(\kappa_1+\kappa_2)}$

where $\kappa$, $\kappa_1$, $\kappa_2 \geq 0$.

**Definition 5[58].** Let $IVIFN = \left(\left[LM, RM\right], \left[LN, RN\right]\right)$ be IVIFN, and the score functions $SF\left(IVIFN\right)$ and $AF\left(IVIFN\right)$ of IVIIFN are defined as follows:

$$SF\left(IVIFN\right) = \left(\frac{\left(LM + RM\right)\left(LM + LN\right) - \left(LN + RN\right)\left(RM + RN\right)}{2}\right) \tag{2.14}$$

where $SF\left(IVIFN\right)\in\left[-1,1\right]$, and the greater the value of $SF\left(IVIFN\right)$ is, the larger the corresponding IVIFN is.

$$AF\left(IVIFN\right)=\frac{\left(1-LM+RM\right)\left(1-LM-LN\right)+\left(1-LN+RN\right)\left(1-RM-RN\right)}{2} \quad (2.15)$$

where $AF\left(IVIFN\right)\in\left[0,1\right]$, and the greater the value of $AF\left(IVIFN\right)$ is, the larger the corresponding IVIFN is.

**Definition 6[58].** According to Definition 4, suppose any two IVIFNs $IVIFN_1=\left(\left[LM_1,RM_1\right],\left[LN_1,RN_1\right]\right)$ and $IVIFN_2=\left(\left[LM_2,RM_2\right],\left[LN_2,RN_2\right]\right)$, and the comparison of these two IVIFNs is described as follows:

If $SF\left(IVIFN_1\right)>SF\left(IVIFN_2\right)$, then $IVIFN_1>IVIFN_2$.

If $SF\left(IVIFN_1\right)<SF\left(IVIFN_2\right)$, then $IVIFN_1<IVIFN_2$.

If $SF\left(IVIFN_1\right)=SF\left(IVIFN_2\right)$, then it can be divided into the following three cases:

If $AF\left(IVIFN_1\right)<AF\left(IVIFN_2\right)$, then $IVIFN_1<IVIFN_2$.

a) If $AF\left(IVIFN_1\right)>AF\left(IVIFN_2\right)$, then $IVIFN_1>IVIFN_2$.

b) If $AF\left(IVIFN_1\right)=AF\left(IVIFN_2\right)$, then $IVIFN_1=IVIFN_2$.

### 2.2 Correlated aggregation operator

**Definition 7[55].** Let a set of n-dimensional IVIFNs $IVIFN_m=\left(\left[LM_m,RM_m\right],\left[LN_m,RN_m\right]\right)$, $\left(m=1,2,\cdots,n\right)$. According to Definition 3, the IVIF weighted average operators (IVIFWA) are defined as follows:

$$\begin{aligned}IVIFWA_{\vartheta}\left(IVIFN_1,IVIFN_2,\cdots,IVIFN_n\right)&=\sum_{m=1}^{n}\vartheta_m IVIFN_m\\&=\left(\left[1-\prod_{k=1}^{n}\left(1-LM_m\right)^{\vartheta_m},1-\prod_{m=1}^{n}\left(1-RM_m\right)^{\vartheta_m}\right],\left[\prod_{m=1}^{n}LN_m^{\vartheta_m},\prod_{m=1}^{n}RN_m^{\vartheta_m}\right]\right)\end{aligned} \quad (2.16)$$

where $\vartheta=\left(\vartheta_1,\vartheta_2,\ldots,\vartheta_n\right)^T$ is the weighting vector of $IVIFN_m\left(m=1,2,\cdots,n\right)$, and $\forall\vartheta_m\in\left[0,1\right]$, $\sum_{m=1}^{n}\vartheta_m=1$.

Specially, when $\vartheta=\left(\frac{1}{n},\frac{1}{n},\ldots,\frac{1}{n}\right)^T$, IVIFWA operators degenerate into IVIF average operators (IVIFA).

**Definition 7[56].** Let a set of n-dimensional IVIFNs $IVIFN_m=\left(\left[LM_m,RM_m\right],\left[LN_m,RN_m\right]\right)$, $\left(m=1,2,\cdots,n\right)$. According to Definition 3, IVIF weighted geometric operators (IVIFWG) operators can be defined as follows:

$$\begin{aligned}&IVIFWG_\vartheta\left(IVIFN_1,IVIFN_2,\cdots,IVIFN_n\right)=\prod_{m=1}^{n}\left(IVIFN_m\right)^{\vartheta_m}\\&=\left(\left[\prod_{m=1}^{n}LM_m^{\vartheta_m},\prod_{m=1}^{n}RM_m^{\vartheta_m}\right],\left[1-\prod_{m=1}^{n}\left(1-LN_m\right)^{\vartheta_m},1-\prod_{m=1}^{n}\left(1-RN_m\right)^{\vartheta_m}\right]\right)\end{aligned}\tag{2.17}$$

where $\vartheta=\left(\vartheta_1,\vartheta_2,\ldots,\vartheta_n\right)^T$ is the weighting vector of $IVIFN_m\left(m=1,2,\cdots,n\right)$, and $\forall\vartheta_m\in\left[0,1\right]$, $\sum_{m=1}^{n}\vartheta_m=1$.

Specially, when $\vartheta=\left(\frac{1}{n},\frac{1}{n},\ldots,\frac{1}{n}\right)^T$, IVIFWG operators degenerate into IVIF geometric operators (IVIFG).

**Definition 8[13, 55].** Suppose two IVIFNs $IIFN_1=\left(\left[LM_1,RM_1\right],\left[LN_1,RN_1\right]\right)$ and $IIFN_2=\left(\left[LM_2,RM_2\right],\left[LN_2,RN_2\right]\right)$, then the Hamming and Hausdorff and its hybrid distance are defined respectively as follows:

$$\begin{aligned}&DHM\left(IVIFN_1,IVIFN_2\right)\\&=\frac{1}{4}\left(\left|LM_1-LM_2\right|+\left|RM_1-RM_2\right|+\left|LN_1-LN_2\right|+\left|RN_1-RN_2\right|\right)\end{aligned}\tag{2.18}$$

$$\begin{aligned}&DHA\left(IVIFN_1,IVIFN_2\right)\\&=\frac{1}{2}\max\left\{\left|LM_1-LM_2\right|,\left|RM_1-RM_2\right|,\left|LN_1-LN_2\right|,\left|RN_1-RN_2\right|\right\}\end{aligned}\tag{2.19}$$

$$\begin{aligned}DM\left(IVIFN_1,IVIFN_2\right)&=\frac{1}{4}\left(\begin{matrix}\left|LM_1-LM_2\right|+&\left|RM_1-RM_2\right|\\+\left|LN_1-LN_2\right|&+\left|RN_1-RN_2\right|\end{matrix}\right)\\&+\frac{1}{2}\max\left\{\begin{matrix}\left|LM_1-LM_2\right|,&\left|RM_1-RM_2\right|\\,\left|LN_1-LN_2\right|&,\left|RN_1-RN_2\right|\end{matrix}\right\}\end{aligned}\tag{2.20}$$

## 3. The extended EDAS method based on CPT with IVIFSs

Keshavarz Ghorabaee, Zavadskas, Olfat and Turskis [47] proposed a new evaluation approach

named as EDAS method which is based on the average distance between solutions in 2015. In this method, the average scheme of alternative schemes under all attributes is selected as the reference point, and then the positive and negative distances between each scheme and the average scheme are calculated respectively. Finally, the optimal scheme is selected by synthesizing the positive and negative distances. In this section, the influence of the DMs' risk attitude on the decision result is considered in the classical EDAS method, and the IVIFN is used to represent the evaluation value of the alternative attribute when the attribute weight is completely unknown. An IVIF-EDAS method based on cumulative prospect theory (CPT-IVIF-EDAS) is constructed to solve MAGDM problem in uncertain environment.

**3.1 The extended EDAS method based on CPT**

In a MADM problem, there are sets of alternatives $HL = \{NL_1, NL_2, \cdots, NL_n\}$, attributes $HT = \{NT_1, NT_2, \cdots, NT_k\}$ and the weighting vector of attributes $\varpi = (\varpi_1, \varpi_2, \cdots, \varpi_k)^T$, and $\forall \varpi_s \in [0,1]$, $\sum_{s=1}^{k} \varpi_s = 1$. Therefore, $y_{rs}\ (r = 1, 2, \cdots, n; s = 1, 2, \cdots, k)$ means the evaluation value of the $r-th$ alternative under the $s-th$ attribute. According to the evaluation values, the MADM decision matrix is constructed as follows:

$$Y = [y_{rs}]_{n\times k} = \begin{bmatrix} y_{11} & y_{12} & \cdots & y_{1s} & \cdots & y_{1k} \\ y_{21} & y_{22} & \cdots & y_{2s} & \cdots & y_{2k} \\ \vdots & \vdots & \ddots & \vdots & \ddots & \vdots \\ y_{r1} & y_{r2} & \cdots & y_{rs} & \cdots & y_{rk} \\ \vdots & \vdots & \ddots & \vdots & \ddots & \vdots \\ y_{n1} & y_{n2} & \cdots & y_{ns} & \cdots & y_{nk} \end{bmatrix} \tag{3.1}$$

The extended EDAS method for MADM based on CPT is given below[59].

**Step 1.** The average solution under each attribute $\Im = [\Im_s]_{1\times k}$ is calculated by Eq. (3.2).

$$\Im_s = \frac{\sum_{r=1}^{n} y_{rs}}{n} \tag{3.2}$$

**Step 2.** Calculate the relative attribute weighting matrix $G = [g_{rs}(\varpi_s)]_{1\times k}$.

Based on the size relationship between $y_{rs}$ and $\Im_s$, the relative attribute weight $g_{rs}(\varpi_s)$ is calculated by Eq. (3.3).

$$g_{rs}\left(\varpi_s\right)=\begin{cases}\dfrac{\varpi_s^{\alpha}}{\left(\varpi_s^{\alpha}+\left(1-\varpi_s\right)^{\alpha}\right)^{\frac{1}{\alpha}}}, & y_{rs}\geq\Im_s\\ \dfrac{\varpi_s^{\beta}}{\left(\varpi_s^{\beta}+\left(1-\varpi_s\right)^{\beta}\right)^{\frac{1}{\beta}}}, & y_{rs}<\Im_s\end{cases}\tag{3.3}$$
$$r=1,2,\cdots,n;s=1,2,\cdots,k$$

where $\alpha,\beta$ is the curvature parameter of the weighting function, and $0<\alpha\leq1,0<\beta\leq1$.

**Step 3.** Calculate the positive and negative ideal distance between average solution $\Im$ (PDA and NDA) and the alternative $HL_r$ under attribute $HT_s$.

According to the type of attribute $HT_s$, the equations (3.4) and (3.5) are constructed by introducing the value function of CPT. Then the $PDA_{rs}$ and $NDA_{rs}$ means the PDA and NDA between alternative $HL_r$ under attribute $HT_s$ and average solution:

When $HT_s$ is positive attribute:

$$\begin{cases}PDA_{rs}=\dfrac{\left(\max(0,(y_{rs}-\Im_s))\right)^{\gamma}}{\Im_s} & ,y_{rs}\geq\Im_s\\ NDA_{rs}=\dfrac{\rho\cdot\left(\max(0,(\Im_s-y_{rs}))\right)^{\delta}}{\Im_s} & ,\ y_{rs}<\Im_s\end{cases}\tag{3.4}$$
$$r=1,2,\cdots,n;s=1,2,\cdots,k$$

When $HT_s$ is negative attribute:

$$\begin{cases}PDA_{rs}=\dfrac{\left(\max(0,(\Im_s-y_{rs}))\right)^{\gamma}}{\Im_s} & ,y_{rs}\geq\Im_s\\ NDA_{rs}=\dfrac{\rho\cdot\left(\max(0,(y_{rs}-\Im_s))\right)^{\delta}}{\Im_s} & ,\ y_{rs}<\Im_s\end{cases}\tag{3.5}$$
$$r=1,2,\cdots,n;s=1,2,\cdots,k$$

where $\gamma\left(0<\gamma\leq1\right)$ and $\delta\left(0<\delta\leq1\right)$ mean the curvature parameters of value function, and $\rho\left(\rho>1\right)$ denotes the sensitivity of DMs to avoid losses.

**Step 4.** Calculate the weighted positive distance $SP_r$ and negative distance $SN_r$.

According to the relative attribute weight $g_{rs}\left(\varpi_s\right)$ by Step 2, $SP_r$ and $SN_r$ are obtained by Eq. (3.6) and (3.7):

$$SP_r = \sum_{s=1}^{e} g_{rs}\left(\varpi_s\right)\cdot PDA_{rs}, r = 1,2,\cdots,n \tag{3.6}$$

$$SN_r = \sum_{s=1}^{e} g_{rs}\left(\varpi_s\right)\cdot NDA_{rs}, r = 1,2,\cdots,n \tag{3.7}$$

**Step 5.** The normalized $SP_r$ and $SN_r$ ( $NSP_r$ and $NSN_r$ ) is calculated by Eq. (3.8) and (3.9).

$$NSP_r = \frac{SP_r}{\max_r\left(SP_r\right)}, r = 1,2,\cdots,n \tag{3.8}$$

$$NSN_r = 1 - \frac{SN_r}{\max_r\left(SN_r\right)}, r = 1,2,\cdots,n \tag{3.8}$$

**Step 6.** The overall assessment score $S_r$ is defined by Eq. (3.10).

$$S_r\left(HL_r\right) = \frac{1}{2}\left(NSP_r + NSN_r\right), r = 1,2,\cdots,n \tag{3.10}$$

**Step 7.** The ranking results of alternatives.

The descending ranking results of all alternatives $HL_r\left(r = 1,2,\cdots,n\right)$ is obtained according to the overall assessment score $S_r\left(r = 1,2,\cdots,n\right)$ calculated by Step 6. The larger the value of overal assessment score $S_r$ is, the better the corresponding alternative is, and it is supposed to be chosen by DMs. Otherwise, it should be abandoned.

### 3.2 The extended EDAS method based on CPT in IVIFSs

According to the related theory of IVIFSs and CPT-EDAS method in Section 3.1, we introduce IVIFN to reconstruct the method to solve MAGDM problem, namely IVIF-CPT-EDAS method. The detailed steps are as follows (Figure. 3.1):

The weighting vector of expert set $HD = \left\{HD_1, HD_2, \cdots, HD_e\right\}$ is $\nu = \left(\nu_1, \nu_2, \cdots, \nu_l\right)^T$ satisfies the condition that $\forall \nu_l \in \left[0,1\right], \sum_{l=1}^{e} \nu_l = 1$. The alternatives set $HL = \left\{HL_1, HL_2, \cdots, HL_n\right\}$ and attributes set $HT = \left\{HT_1, HT_2, \cdots, HT_k\right\}$ and the attribute weighting vector $\varpi = \left(\varpi_1, \varpi_2, \cdots, \varpi_k\right)^T$, and $\forall \varpi_s \in \left[0,1\right], \sum_{s=1}^{k} \varpi_s = 1$. Therefore, the assessment value of the of the $r-th$ alternative under the $s-th$ attribute by $l-th$ expert is denoted by IVIFN:

$\Re_{rs}^{(l)} = \left(\left[LM_{rs}^{(l)}, RM_{rs}^{(l)}\right], \left[LN_{rs}^{(l)}, RN_{rs}^{(l)}\right]\right)$, where $\left[LM_{rs}^{(l)}, RM_{rs}^{(l)}\right]$ and $\left[LN_{rs}^{(l)}, RN_{rs}^{(l)}\right]$ $\left(r = 1, 2, \cdots, n; s = 1, 2, \cdots, k; l = 1, 2, \cdots, e\right)$ denote the membership and non-membership interval.

By the previously mentioned, the IVIF MADM decision matrix is constructed by $n \times k$ IVIF evaluation value:

$$R^{(l)} = \left[\Re_{rs}^{(l)}\right]_{n\times k} = \begin{bmatrix} \Re_{11}^{(l)} & \Re_{12}^{(l)} & \cdots & \Re_{1s}^{(l)} & \cdots & \Re_{1k}^{(l)} \\ \Re_{21}^{(l)} & \Re_{22}^{(l)} & \cdots & \Re_{2s}^{(l)} & \cdots & \Re_{2k}^{(l)} \\ \vdots & \vdots & \ddots & \vdots & \ddots & \vdots \\ \Re_{r1}^{(l)} & \Re_{r2}^{(l)} & \cdots & \Re_{rs}^{(l)} & \cdots & \Re_{rk}^{(l)} \\ \vdots & \vdots & \ddots & \vdots & \ddots & \vdots \\ \Re_{n1}^{(l)} & \Re_{n2}^{(l)} & \cdots & \Re_{ns}^{(l)} & \cdots & \Re_{nk}^{(l)} \end{bmatrix} \tag{3.6}$$

$$r = 1, 2, \cdots, n; s = 1, 2, \cdots, k; l = 1, 2, \cdots e$$

**Step 1.** Obtain the IVIF decision matrix $\aleph$.

Integrate IVIF decision matrix $R^{(l)}\left(l = 1, 2, \cdots, e\right)$ by IVIFWA operators, and obtain the IVIF decision matrix $\aleph = \left[\Re_{rs}\right]_{n\times k}$ using Eq. (3.7).

$$\begin{aligned} \Re_{rs} &= \left(\left[LM_{rs}, RM_{rs}\right], \left[LN_{rs}, RN_{rs}\right]\right) = IIFWA_{\vartheta}\left(\Re_{rs}^{(1)}, \Re_{rs}^{(2)}, \cdots, \Re_{rs}^{(l)}\right) \\ &= \left(\left[1 - \prod_{l=1}^{e}\left(1 - LM_{rs}^{(l)}\right)^{v_l}, 1 - \prod_{l=1}^{e}\left(1 - RM_{rs}^{(l)}\right)^{v_l}\right], \left[\prod_{l=1}^{e}\left(LN_{rs}^{(l)}\right)^{v_l}, \prod_{l=1}^{e}\left(RN_{rs}^{(l)}\right)^{v_l}\right]\right) \end{aligned} \tag{3.7}$$

**Step 2.** Calculate the normalized IVIF decision matrix $\aleph^*$.

Transform the IVIF decision matrix $\aleph = \left[\Re_{rs}\right]_{n\times k}$ into normalized IVIF decision matrix $\aleph^* = \left[\Re_{rs}^*\right]_{n\times k} = \left(\left[LM_{rs}^*, RM_{rs}^*\right], \left[LN_{rs}^*, RN_{rs}^*\right]\right)_{n\times k}$ using Eq. (3.8).

$$\Re_{rs}^* = \begin{cases} \left(\left[LM_{rs}, RM_{rs}\right], \left[LN_{rs}, RN_{rs}\right]\right), & attribute\ HT_s\ \ is\ \ positive \\ \left(\left[LN_{rs}, RN_{rs}\right], \left[LM_{rs}, RM_{rs}\right]\right), & attribute\ HT_s\ \ is\ \ negative \end{cases} \tag{3.8}$$

$$\mathrm{r} = 1, 2, \cdots, n; s = 1, 2, \cdots, k$$

**Step 3.** Determine the original attribute weight $\varpi_s\left(s = 1, 2, \cdots, k\right)$ using extended entropy method. the specific steps of IVIF-entropy method are shown as follows:

(1) Determine the IVIF negative ideal point (IVIF-NIP): $\Re_s^{*-}\left(s = 1, 2, \cdots, k\right)$, and satisfying the below condition:

$$\begin{aligned}\overset{*-}{\Re}_s &= \left(\left[\min_r LM^*_{rs}, \min_r RM^*_{rs}\right], \left[\max_r LN^*_{rs}, \max_r RN^*_{rs}\right]\right) \\ &= \left(\left[LM^{*-}_{rs}, RM^{*-}_{rs}\right], \left[LN^{*-}_{rs}, RN^{*-}_{rs}\right]\right)\end{aligned} \tag{3.9}$$

(2) Obtain the hybrid distance matrix $\Lambda = \left[DM_{rs}\left(\overset{*}{\Re}_{rs}, \overset{*-}{\Re}_s\right)\right]_{n\times k}$ using Eq. (3.10)

$$\begin{aligned}DM_{rs}\left(\overset{*}{\Re}_{rs}, \overset{*-}{\Re}_s\right) &= \frac{1}{4}\left(\begin{array}{ll}\left|LM^*_{rs} - LM^{*-}_{rs}\right| + & \left|RM^*_{rs} - RM^{*-}_{rs}\right| \\ +\left|LN^*_{rs} - LN^{*-}_{rs}\right| & +\left|RN^*_{rs} - RN^{*-}_{rs}\right|\end{array}\right) \\ &+ \frac{1}{2}\max\left\{\begin{array}{ll}\left|LM^*_{rs} - LM^{*-}_{rs}\right|, & \left|RM^*_{rs} - RM^{*-}_{rs}\right| \\ ,\left|LN^*_{rs} - LN^{*-}_{rs}\right| & ,\left|RN^*_{rs} - RN^{*-}_{rs}\right|\end{array}\right\} \\ & r = 1, 2, \cdots n; s = 1, 2, \cdots, k\end{aligned} \tag{3.10}$$

(3) Calculate the normalized distance matrix $\Lambda' = \left[DM'_{rs}\left(\overset{*}{\Re}_{rs}, \overset{*-}{\Re}_s\right)\right]_{n\times k}$ using Eq. (3.11)

$$DM'_{rs}\left(\overset{*}{\Re}_{rs}, \overset{*-}{\Re}_s\right) = \frac{DM_{rs}\left(\overset{*}{\Re}_{rs}, \overset{*-}{\Re}_s\right)}{\sum_{r=1}^{n} DM_{rs}\left(\overset{*}{\Re}_{rs}, \overset{*-}{\Re}_s\right)}, \ r = 1, 2, \cdots, n; s = 1, 2, \cdots, k \tag{3.11}$$

(4) Calculate the entropy degree $E_s$ of the $s\text{-}th$ attribute $HT_s$ using Eq. (3.12).

$$\begin{aligned}E_s &= -\frac{1}{\ln n}\sum_{r=1}^{n} DM'_{rs}\left(\overset{*}{\Re}_{rs}, \overset{*-}{\Re}_s\right) \cdot \ln DM'_{rs}\left(\overset{*}{\Re}_{rs}, \overset{*-}{\Re}_s\right) \\ & s = 1, 2, \cdots, k; 0 \le E_s \le 1\end{aligned} \tag{3.12}$$

(5) Figure out the original attribute weight $\varpi_s$ using Eq. (3.13).

$$\varpi_s = \frac{1 - E_s}{\sum_{s=1}^{k} 1 - E_s}, s = 1, 2, \cdots k \tag{3.13}$$

**Step 4.** Determine the average solution $AV_s\ (s = 1, 2, \cdots, k)$ using Eq. (3.14) with the information of IVIFWA operators and normalized IVIF decision matrix $\aleph^*$.

$$AV_s = \left(\begin{array}{l}\left[1 - \left(\prod_{r=1}^{n}\left(1 - LM^*_{rs}\right)\right)^{\frac{1}{n}}, 1 - \left(\prod_{r=1}^{n}\left(1 - RM^*_{rs}\right)\right)^{\frac{1}{n}}\right] \\ , \left[\left(\prod_{r=1}^{n} LN^*_{rs}\right)^{\frac{1}{n}}, \left(\prod_{r=1}^{n} RN^*_{rs}\right)^{\frac{1}{n}}\right]\end{array}\right) \tag{3.14}$$

**Step 5.** Calculate the relative attribute matrix $G' = \left[g'_s\left(\varpi_s\right)\right]_{n\times k}$.

Calculate the relative attribute weight $g'_{rs}\left(\varpi_s\right)$ using Eq. (3.15).

$$g'_{rs}\left(\varpi_s\right)=\begin{cases}\dfrac{\varpi_s^{\alpha}}{\left(\varpi_s^{\alpha}+\left(1-\varpi_s\right)^{\alpha}\right)^{\frac{1}{\alpha}}}, & \overset{*}{\Re}_{rs}\geq AV_s\\ \dfrac{\varpi_s^{\beta}}{\left(\varpi_s^{\beta}+\left(1-\varpi_s\right)^{\beta}\right)^{\frac{1}{\beta}}}, & \overset{*}{\Re}_{rs}<AV_s\end{cases} \tag{3.15}$$

$$r=1,2,\cdots,n;s=1,2,\cdots,k$$

**Step 6.** Calculate the $PDA'_{rs}$ and $NDA'_{rs}$.

$$PDA'_{rs}=\begin{cases}\dfrac{\left(DM(\overset{*}{\Re}_{rs},AV_s)\right)^{\gamma}}{AF\left(AV_s\right)} & ,\overset{*}{\Re}_{rs}\geq AV_s\\ 0 & ,\overset{*}{\Re}_{rs}<AV_s\end{cases} \tag{3.16}$$

$$r=1,2,\cdots,n;s=1,2,\cdots,k$$

$$NDA'_{rs}=\begin{cases}0 & ,\overset{*}{\Re}_{rs}\geq AV_s\\ \dfrac{\rho\cdot\left(DM(AV_s,\overset{*}{\Re}_{rs})\right)^{\delta}}{AF\left(AV_s\right)} & ,\overset{*}{\Re}_{rs}<AV_s\end{cases} \tag{3.17}$$

$$r=1,2,\cdots,n;s=1,2,\cdots,k$$

**Step 7.** Calculate the weighted positive and negative distance ($SP'_r$ and $SN'_r$) using Eq. (3.18) and (3.19).

$$SP'_r=\sum_{s=1}^{k}g'_{rs}\left(\varpi_s\right)\cdot PDA'_{rs},\ r=1,2,\cdots,n \tag{3.18}$$

$$SN'_r=\sum_{s=1}^{k}g'_{rs}\left(\varpi_s\right)\cdot NDA'_{rs},\ r=1,2,\cdots,n \tag{3.19}$$

**Step 8.** Calculate the normalized weighted positive and negative distance ($NSP'_r$ and $NSN'_r$).

Similar to Eq. (3.8) and (3.9), Eq. (3.20) and (3.21) are used to standardize the weighted positive and negative distance of the alternatives, and the normalized weighted positive and negative distance ($NSP'_r$ and $NSN'_r$) are obtained by Eq. (3.20) and (3.21).

$$NSP'_r=\frac{SP'_r}{\max\limits_r\left(SP'_r\right)},r=1,2,\cdots,n \tag{3.20}$$

$$NSN'_r = 1 - \frac{SN'_r}{\max_r \left( SN'_r \right)}, r = 1, 2, \cdots, n \tag{3.21}$$

**Step 9.** The overall assessment score $S'_r$ of each alternative $HL_r$ using Eq. (3.22).

$$S'_r \left( HL_r \right) = \frac{1}{2} \left( NSP'_r + NSN'_r \right), r = 1, 2, \cdots, n \tag{3.22}$$

**Step 10.** Based on the overall assessment score $S'_r \left( r = 1, 2, \cdots, n \right)$ in Step 9, rank all the alternatives in descending order. The larger the value of overall assessment score $S'_r$ is, the better the alternative is.

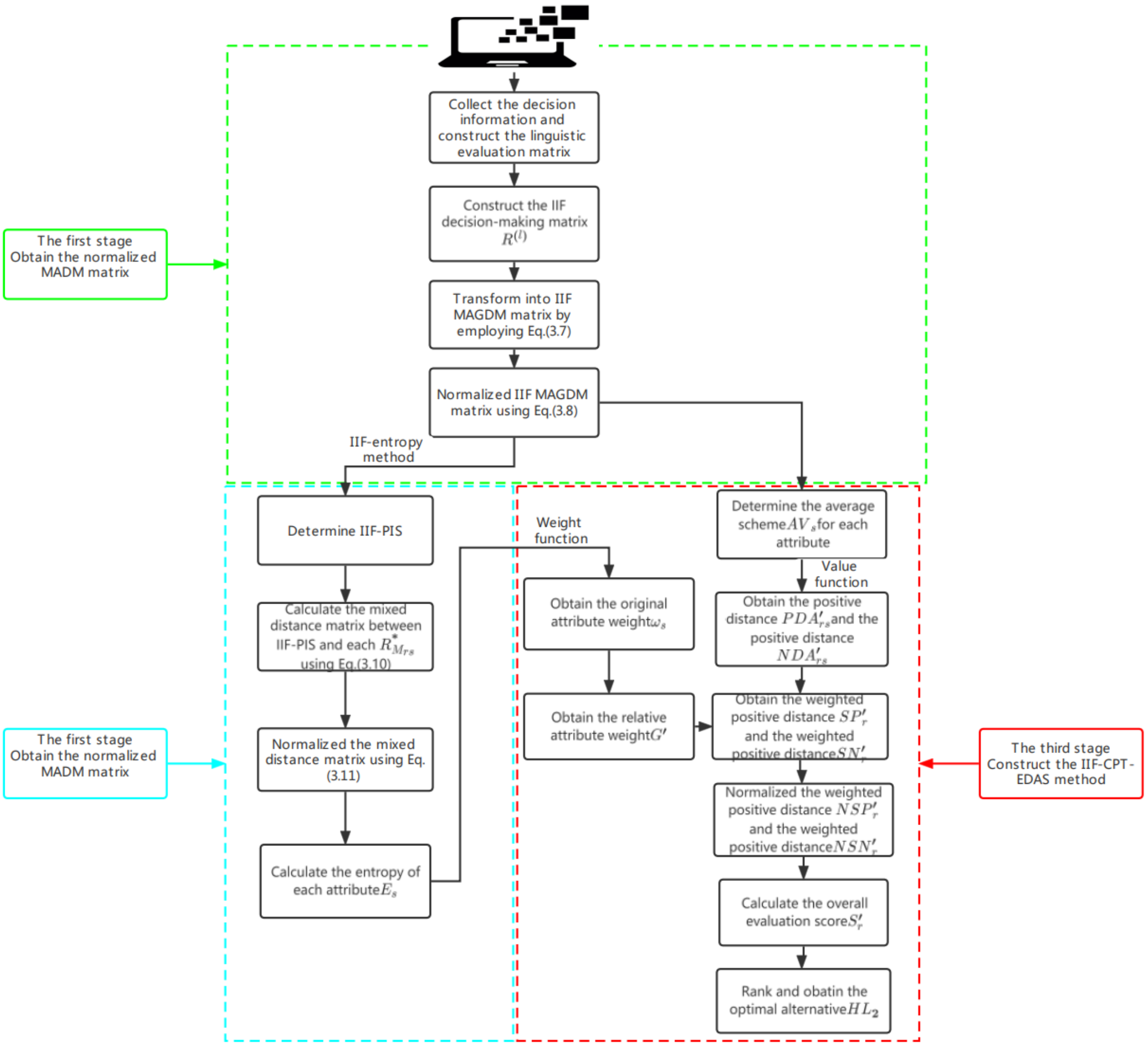

Figure 3.1 the flow chart of IVIF-CPT-EDAS

## 4. Application of IVIF MAGDM decision model based on CPT in green technology venture capital

### 4.1 Background Description ·

Development is always an eternal theme. However, due to the limitation and irreversibility of resources, in order to reduce pollution, save resources, take the road of sustainable development, and realize the harmonious development of man and nature, the development of green economy has become a new demand. From the perspective of investment, as one of the five key green investment areas, green technology venture capital has a broad development space to promote the development of circular economy, such as comprehensive utilization of resources technology, environmental protection technology, ecological agriculture technology, comprehensive utilization of resources technology, new energy technology and so on. At the same time, as a DM, in the face of many new potential projects, it becomes particularly important to scientifically and quickly evaluate the investment projects that obtain significant benefits. Therefore, this paper takes the project selection of green technology venture capital as an example to verify the effectiveness and applicability of the improved EDAS method.

**4.2 Background description**

We invite five experts $HD_c\left(c=1,2,\cdots,5\right)$ to select the best solution from five green technology venture capital projects $HL_r\left(r=1,2,\cdots,5\right)$ under six attributes $HT_s\left(s=1,2,\cdots,6\right)$. The specific description of attributes is as follows: $HT_1$: return on investment; $HT_2$: amount of waste produced; $HT_3$: resources are renewable; $HT_4$: difficulty of capital exit; $HT_5$: degree of market demand; $HT_6$: The R&D capability of the technical research team, where $HT_2$ and $HT_4$ are the cost attributes， and others are benefit attributes. Meanwhile, the weighting vector $\nu=\left(\nu_1,\nu_2,\cdots,\nu_5\right)=\left(0.29,0.17,0.19,0.15,0.20\right)$ of five experts is given.

Table 4.1 IVIF linguistic evaluation scale

| Linguistic scale | IVIFN |
|---|---|
| Extremely terrible (ET) | $\langle[0.00,0.10],[0.85,0.90]\rangle$ |
| Very terrible (VT) | $\langle[0.00,0.10],[0.70,0.75]\rangle$ |
| Terrible (T) | $\langle[0.15,0.25],[0.55,0.60]\rangle$ |
| Middle terrible (MT) | $\langle[0.30,0.40],[0.45,0.50]\rangle$ |
| Middle (M) | $\langle[0.40,0.50],[0.35,0.40]\rangle$ |
| Middle good (MG) | $\langle [0.50, 0.60], [0.25, 0.30] \rangle$ |
| Good (G) | $\langle [0.60, 0.70], [0.15, 0.20] \rangle$ |
| Very good (VG) | $\langle [0.70, 0.80], [0.05, 0.10] \rangle$ |
| Extremely good (EG) | $\langle [0.80, 0.90], [0.05, 0.10] \rangle$ |
| Perfectly good (PG) | $\langle [1.00, 1.00], [0.00, 0.00] \rangle$ |

Table 4.2 the linguistic assessment matrix $M_1$ by the first expert

| | $HT_1$ | $HT_2$ | $HT_3$ | $HT_4$ | $HT_5$ | $HT_6$ |
|---|---|---|---|---|---|---|
| $HL_1$ | MG | G | M | G | MG | MG |
| $HL_2$ | VG | M | EG | M | EG | G |
| $HL_3$ | MG | VG | M | MG | G | M |
| $HL_4$ | M | G | MT | M | MG | T |
| $HL_5$ | G | VT | M | M | G | MG |

Table 4.3 the linguistic assessment matrix $M_2$ by the second expert

| | $HT_1$ | $HT_2$ | $HT_3$ | $HT_4$ | $HT_5$ | $HT_6$ |
|---|---|---|---|---|---|---|
| $HL_1$ | G | MG | MG | G | M | MG |
| $HL_2$ | EG | MT | EG | M | EG | G |
| $HL_3$ | G | VG | M | M | MG | MT |
| $HL_4$ | MT | G | M | M | G | T |
| $HL_5$ | MG | MT | MT | MT | MG | M |

Table 4.4 the linguistic assessment matrix $M_3$ by the third expert

| | $HT_1$ | $HT_2$ | $HT_3$ | $HT_4$ | $HT_5$ | $HT_6$ |
|---|---|---|---|---|---|---|
| $HL_1$ | MG | G | G | MG | G | M |
| $HL_2$ | EG | M | VG | G | EG | G |
| $HL_3$ | MG | G | MG | MG | M | MT |
| $HL_4$ | M | G | M | MG | G | MT |
| $HL_5$ | G | M | VT | M | MG | MG |

Table 4.5 the linguistic assessment matrix $M_4$ by the fourth expert

| | $HT_1$ | $HT_2$ | $HT_3$ | $HT_4$ | $HT_5$ | $HT_6$ |
|---|---|---|---|---|---|---|
| $HL_1$ | G | G | M | G | G | MG |
| $HL_2$ | EG | M | EG | M | VG | VG |
| $HL_3$ | M | EG | MG | MG | MG | M |
| $HL_4$ | MT | G | MG | M | G | MT |
| $HL_5$ | G | MT | MT | M | MG | M |

Table 4.6 the linguistic assessment matrix $M_5$ by the fifth expert

| | $HT_1$ | $HT_2$ | $HT_3$ | $HT_4$ | $HT_5$ | $HT_6$ |
|---|---|---|---|---|---|---|
| $HL_1$ | VG | MG | G | G | MG | G |
| $HL_2$ | EG | MT | EG | MG | VG | VG |
| $HL_3$ | MG | VG | G | MG | G | M |
| $HL_4$ | M | G | MG | M | VG | MT |
| $HL_5$ | VG | M | MT | M | M | MG |

Table 4.7 the IVIF decision matrix $R^{(1)}$ by the first expert

| | $HT_1$ | $HT_2$ | $HT_3$ |
|---|---|---|---|
| $HL_1$ | $\langle[0.50,0.60],[0.25,0.30]\rangle$ | $\langle[0.60,0.70],[0.15,0.20]\rangle$ | $\langle[0.40,0.50],[0.35,0.40]\rangle$ |
| $HL_2$ | $\langle[0.70,0.80],[0.05,0.10]\rangle$ | $\langle[0.40,0.50],[0.35,0.40]\rangle$ | $\langle[0.80,0.90],[0.05,0.10]\rangle$ |
| $HL_3$ | $\langle[0.50,0.60],[0.25,0.30]\rangle$ | $\langle[0.70,0.80],[0.05,0.10]\rangle$ | $\langle[0.40,0.50],[0.35,0.40]\rangle$ |
| $HL_4$ | $\langle[0.40,0.50],[0.35,0.40]\rangle$ | $\langle[0.60,0.70],[0.15,0.20]\rangle$ | $\langle[0.30,0.40],[0.45,0.50]\rangle$ |
| $HL_5$ | $\langle[0.60,0.70],[0.15,0.20]\rangle$ | $\langle[0.00,0.10],[0.70,0.75]\rangle$ | $\langle[0.40,0.50],[0.35,0.40]\rangle$ |
| | $HT_4$ | $HT_5$ | $HT_6$ |
| $HL_1$ | $\langle[0.60,0.70],[0.15,0.20]\rangle$ | $\langle[0.50,0.60],[0.25,0.30]\rangle$ | $\langle[0.50,0.60],[0.25,0.30]\rangle$ |
| $HL_2$ | $\langle[0.40,0.50],[0.35,0.40]\rangle$ | $\langle[0.80,0.90],[0.05,0.10]\rangle$ | $\langle[0.60,0.70],[0.15,0.20]\rangle$ |
| $HL_3$ | $\langle[0.50,0.60],[0.25,0.30]\rangle$ | $\langle[0.60,0.70],[0.15,0.20]\rangle$ | $\langle[0.40,0.50],[0.35,0.40]\rangle$ |
| $HL_4$ | $\langle[0.40,0.50],[0.35,0.40]\rangle$ | $\langle[0.50,0.60],[0.25,0.30]\rangle$ | $\langle[0.15,0.25],[0.55,0.60]\rangle$ |
| $HL_5$ | $\langle[0.40,0.50],[0.35,0.40]\rangle$ | $\langle[0.60,0.70],[0.15,0.20]\rangle$ | $\langle[0.50,0.60],[0.25,0.30]\rangle$ |

Table 4.8 the IVIF decision matrix $R^{(2)}$ by the second expert

| | $HT_1$ | $HT_2$ | $HT_3$ |
|---|---|---|---|
| $HL_1$ | $\langle[0.60,0.70],[0.15,0.20]\rangle$ | $\langle[0.50,0.60],[0.25,0.30]\rangle$ | $\langle[0.50,0.60],[0.25,0.30]\rangle$ |
| $HL_2$ | $\langle[0.80,0.90],[0.05,0.10]\rangle$ | $\langle[0.30,0.40],[0.45,0.50]\rangle$ | $\langle[0.80,0.90],[0.05,0.10]\rangle$ |
| $HL_3$ | $\langle[0.60,0.70],[0.15,0.20]\rangle$ | $\langle[0.70,0.80],[0.05,0.10]\rangle$ | $\langle[0.40,0.50],[0.35,0.40]\rangle$ |
| $HL_4$ | $\langle[0.30,0.40],[0.45,0.50]\rangle$ | $\langle[0.60,0.70],[0.15,0.20]\rangle$ | $\langle[0.40,0.50],[0.35,0.40]\rangle$ |
| $HL_5$ | $\langle[0.50,0.60],[0.25,0.30]\rangle$ | $\langle[0.30,0.40],[0.45,0.50]\rangle$ | $\langle[0.30,0.40],[0.45,0.50]\rangle$ |

| | $HT_4$ | $HT_5$ | $HT_6$ |
|---|---|---|---|
| $HL_1$ | $\langle[0.60,0.70],[0.15,0.20]\rangle$ | $\langle[0.40,0.50],[0.35,0.40]\rangle$ | $\langle[0.50,0.60],[0.25,0.30]\rangle$ |
| $HL_2$ | $\langle[0.40,0.50],[0.35,0.40]\rangle$ | $\langle[0.80,0.90],[0.05,0.10]\rangle$ | $\langle[0.60,0.70],[0.15,0.20]\rangle$ |
| $HL_3$ | $\langle[0.40,0.50],[0.35,0.40]\rangle$ | $\langle[0.50,0.60],[0.25,0.30]\rangle$ | $\langle[0.30,0.40],[0.45,0.50]\rangle$ |
| $HL_4$ | $\langle[0.40,0.50],[0.35,0.40]\rangle$ | $\langle[0.60,0.70],[0.15,0.20]\rangle$ | $\langle[0.15,0.25],[0.55,0.60]\rangle$ |
| $HL_5$ | $\langle[0.30,0.40],[0.45,0.50]\rangle$ | $\langle[0.50,0.60],[0.25,0.30]\rangle$ | $\langle[0.40,0.50],[0.35,0.40]\rangle$ |

Table 4.9 the IVIF decision matrix $R^{(3)}$ by the third expert

| | $HT_1$ | $HT_2$ | $HT_3$ |
|---|---|---|---|
| $HL_1$ | $\langle[0.50,0.60],[0.25,0.30]\rangle$ | $\langle[0.60,0.70],[0.15,0.20]\rangle$ | $\langle[0.60,0.70],[0.15,0.20]\rangle$ |
| $HL_2$ | $\langle[0.80,0.90],[0.05,0.10]\rangle$ | $\langle[0.40,0.50],[0.35,0.40]\rangle$ | $\langle[0.70,0.80],[0.05,0.10]\rangle$ |
| $HL_3$ | $\langle[0.50,0.60],[0.25,0.30]\rangle$ | $\langle[0.60,0.70],[0.15,0.20]\rangle$ | $\langle[0.50,0.60],[0.25,0.30]\rangle$ |
| $HL_4$ | $\langle[0.40,0.50],[0.35,0.40]\rangle$ | $\langle[0.50,0.60],[0.25,0.30]\rangle$ | $\langle[0.40,0.50],[0.35,0.40]\rangle$ |
| $HL_5$ | $\langle[0.70,0.80],[0.05,0.10]\rangle$ | $\langle[0.40,0.50],[0.35,0.40]\rangle$ | $\langle[0.00,0.10],[0.70,0.75]\rangle$ |
| | $HT_4$ | $HT_5$ | $HT_6$ |
| $HL_1$ | $\langle[0.50,0.60],[0.25,0.30]\rangle$ | $\langle[0.60,0.70],[0.15,0.20]\rangle$ | $\langle[0.40,0.50],[0.35,0.40]\rangle$ |
| $HL_2$ | $\langle[0.60,0.70],[0.15,0.20]\rangle$ | $\langle[0.80,0.90],[0.05,0.10]\rangle$ | $\langle[0.60,0.70],[0.15,0.20]\rangle$ |
| $HL_3$ | $\langle[0.50,0.60],[0.25,0.30]\rangle$ | $\langle[0.60,0.70],[0.15,0.20]\rangle$ | $\langle[0.30,0.40],[0.45,0.50]\rangle$ |
| $HL_4$ | $\langle[0.50,0.60],[0.25,0.30]\rangle$ | $\langle[0.60,0.70],[0.15,0.20]\rangle$ | $\langle[0.30,0.40],[0.45,0.50]\rangle$ |
| $HL_5$ | $\langle[0.40,0.50],[0.35,0.40]\rangle$ | $\langle[0.50,0.60],[0.25,0.30]\rangle$ | $\langle[0.50,0.60],[0.25,0.30]\rangle$ |

Table 4.10 the IVIF decision matrix $R^{(4)}$ by the fourth expert

| | $HT_1$ | $HT_2$ | $HT_3$ |
|---|---|---|---|
| $HL_1$ | $\langle[0.60,0.70],[0.15,0.20]\rangle$ | $\langle[0.60,0.70],[0.15,0.20]\rangle$ | $\langle[0.40,0.50],[0.35,0.40]\rangle$ |
| $HL_2$ | $\langle[0.80,0.90],[0.05,0.10]\rangle$ | $\langle[0.60,0.70],[0.15,0.20]\rangle$ | $\langle[0.80,0.90],[0.05,0.10]\rangle$ |
| $HL_3$ | $\langle[0.40,0.50],[0.35,0.40]\rangle$ | $\langle[0.80,0.90],[0.05,0.10]\rangle$ | $\langle[0.50,0.60],[0.25,0.30]\rangle$ |
| $HL_4$ | $\langle[0.30,0.40],[0.45,0.50]\rangle$ | $\langle[0.60,0.70],[0.15,0.20]\rangle$ | $\langle[0.50,0.60],[0.25,0.30]\rangle$ |
| $HL_5$ | $\langle[0.60,0.70],[0.15,0.20]\rangle$ | $\langle[0.30,0.40],[0.45,0.50]\rangle$ | $\langle[0.30,0.40],[0.45,0.50]\rangle$ |
| | $HT_4$ | $HT_5$ | $HT_6$ |
| $HL_1$ | $\langle[0.60,0.70],[0.15,0.20]\rangle$ | $\langle[0.60,0.70],[0.15,0.20]\rangle$ | $\langle[0.50,0.60],[0.25,0.30]\rangle$ |
| $HL_2$ | $\langle[0.50,0.60],[0.25,0.30]\rangle$ | $\langle[0.70,0.80],[0.05,0.10]\rangle$ | $\langle[0.70,0.80],[0.05,0.10]\rangle$ |
| $HL_3$ | $\langle[0.50,0.60],[0.25,0.30]\rangle$ | $\langle[0.50,0.60],[0.25,0.30]\rangle$ | $\langle[0.40,0.50],[0.35,0.40]\rangle$ |
| $HL_4$ | $\langle[0.40,0.50],[0.35,0.40]\rangle$ | $\langle[0.60,0.70],[0.15,0.20]\rangle$ | $\langle[0.30,0.40],[0.45,0.50]\rangle$ |
| $HL_5$ | $\langle[0.40,0.50],[0.35,0.40]\rangle$ | $\langle[0.50,0.60],[0.25,0.30]\rangle$ | $\langle[0.40,0.50],[0.35,0.40]\rangle$ |

Table 4.11 the IVIF decision matrix $R^{(5)}$ by the fifth expert

| | $HT_1$ | $HT_2$ | $HT_3$ |
|---|---|---|---|
| $HL_1$ | $\langle[0.70,0.80],[0.05,0.10]\rangle$ | $\langle[0.50,0.60],[0.25,0.30]\rangle$ | $\langle[0.60,0.70],[0.15,0.20]\rangle$ |
| $HL_2$ | $\langle[0.80,0.90],[0.05,0.10]\rangle$ | $\langle[0.30,0.40],[0.45,0.50]\rangle$ | $\langle[0.80,0.90],[0.05,0.10]\rangle$ |
| $HL_3$ | $\langle[0.50,0.60],[0.25,0.30]\rangle$ | $\langle[0.70,0.80],[0.05,0.10]\rangle$ | $\langle[0.60,0.70],[0.15,0.20]\rangle$ |
| $HL_4$ | $\langle[0.40,0.50],[0.35,0.40]\rangle$ | $\langle[0.60,0.70],[0.15,0.20]\rangle$ | $\langle[0.50,0.60],[0.25,0.30]\rangle$ |
| $HL_5$ | $\langle[0.70,0.80],[0.05,0.10]\rangle$ | $\langle[0.40,0.50],[0.35,0.40]\rangle$ | $\langle[0.30,0.40],[0.45,0.50]\rangle$ |
| | $HT_4$ | $HT_5$ | $HT_6$ |
| $HL_1$ | $\langle[0.60,0.70],[0.15,0.20]\rangle$ | $\langle[0.50,0.60],[0.25,0.30]\rangle$ | $\langle[0.60,0.70],[0.15,0.20]\rangle$ |
| $HL_2$ | $\langle[0.50,0.60],[0.25,0.30]\rangle$ | $\langle[0.70,0.80],[0.05,0.10]\rangle$ | $\langle[0.70,0.80],[0.05,0.10]\rangle$ |
| $HL_3$ | $\langle[0.50,0.60],[0.25,0.30]\rangle$ | $\langle[0.60,0.70],[0.15,0.20]\rangle$ | $\langle[0.40,0.50],[0.35,0.40]\rangle$ |
| $HL_4$ | $\langle[0.40,0.50],[0.35,0.40]\rangle$ | $\langle[0.70,0.80],[0.05,0.10]\rangle$ | $\langle[0.30,0.40],[0.45,0.50]\rangle$ |
| $HL_5$ | $\langle[0.40,0.50],[0.35,0.40]\rangle$ | $\langle[0.40,0.50],[0.35,0.40]\rangle$ | $\langle[0.50,0.60],[0.25,0.30]\rangle$ |

### 4.2 The example analysis

In this section, the IVIF-CPT-EDAS model was given to descript the process of selecting the most appropriate solution.

**Step 1.** Aggregate the above five IVIF decision matrices $R^{(l)}\left(l=1,2,\cdots,5\right)$ into one IVIF group decision matrix $\aleph$ by using Eq. (3.7).

**Step 2. Obtain the normalized IVIF decision matrix** $\aleph^*$ using Eq. (3.8) in Table 4.12.

Table 4.12 the normalized IVIF decision matrix $\aleph^*$

| | $HT_1$ | $HT_2$ | $HT_3$ |
|---|---|---|---|
| $HL_1$ | $\langle[0.580,0.682],[0.154,0.212]\rangle$ | $\langle[0.181,0.232],[0.566,0.666]\rangle$ | $\langle[0.503,0.606],[0.238,0.291]\rangle$ |
| $HL_2$ | $\langle[0.775,0.878],[0.050,0.100]\rangle$ | $\langle[0.338,0.392],[0.402,0.505]\rangle$ | $\langle[0.784,0.886],[0.050,0.100]\rangle$ |
| $HL_3$ | $\langle[0.505,0.606],[0.241,0.292]\rangle$ | $\langle[0.062,0.114],[0.702,0.805]\rangle$ | $\langle[0.480,0.582],[0.264,0.316]\rangle$ |
| $HL_4$ | $\langle[0.370,0.470],[0.379,0.430]\rangle$ | $\langle[0.165,0.216],[0.583,0.683]\rangle$ | $\langle[0.411,0.512],[0.335,0.386]\rangle$ |
| $HL_5$ | $\langle[0.682,0.788],[0.081,0.136]\rangle$ | $\langle[0.464,0.516],[0.270,0.404]\rangle$ | $\langle[0.284,0.385],[0.455,0.506]\rangle$ |
| | $HT_4$ | $HT_5$ | $HT_6$ |
| $HL_1$ | $\langle[0.165,0.216],[0.583,0.683]\rangle$ | $\langle[0.522,0.623],[0.223,0.274]\rangle$ | $\langle[0.505,0.606],[0.241,0.292]\rangle$ |
| $HL_2$ | $\langle[0.265,0.317],[0.479,0.580]\rangle$ | $\langle[0.770,0.873],[0.050,0.100]\rangle$ | $\langle[0.638,0.740],[0.102,0.157]\rangle$ |
| $HL_3$ | $\langle[0.265,0.315],[0.484,0.585]\rangle$ | $\langle[0.570,0.671],[0.177,0.228]\rangle$ | $\langle[0.366,0.466],[0.383,0.433]\rangle$ |
| $HL_4$ | $\langle[0.328,0.379],[0.420,0.521]\rangle$ | $\langle[0.597,0.699],[0.140,0.196]\rangle$ | $\langle[0.235,0.335],[0.494,0.544]\rangle$ |
| $HL_5$ | $\langle[0.365, 0.415], [0.384, 0.484]\rangle$ | $\langle[0.514, 0.615], [0.231, 0.283]\rangle$ | $\langle[0.470, 0.570], [0.278, 0.329]\rangle$ |

**Step 3.** Determine the original relative attribute weight using $\varpi_s\ (s = 1, 2, \cdots, k)$ extended entropy method.

(1) Determine the IVIF negative point (IVIF-NIS): $\overset{*-}{\Re_s}\ (s = 1, 2, \cdots, k)$ in Eq. (4.13).

Table 4.13 IVIF-NIS $\overset{*-}{\Re_s}$

| IVIF-NIP | $HT_1$ | $HT_2$ | $HT_3$ |
|---|---|---|---|
| $\overset{*-}{\Re_s}$ | $\langle[0.370, 0.470], [0.379, 0.430]\rangle$ | $\langle[0.062, 0.114], [0.702, 0.805]\rangle$ | $\langle[0.284, 0.385], [0.455, 0.506]\rangle$ |
| IVIF-NIP | $HT_4$ | $HT_5$ | $HT_6$ |
| $\overset{*-}{\Re_s}$ | $\langle[0.165, 0.216], [0.583, 0.683]\rangle$ | $\langle[0.514, 0.615], [0.231, 0.283]\rangle$ | $\langle[0.235, 0.335], [0.494, 0.544]\rangle$ |

(2) Calculate the Hamming distance between the IVIF-NIP and all alternatives $HL_r$ under each attribute $HT_s$ matrix $\Lambda = \left[DM_{rs}\left(\overset{*}{\Re}_{rs}, \overset{*-}{\Re_s}\right)\right]_{n\times k}$ using Eq. (3.10).

(3) Calculate the normalized distance matrix $\Lambda' = \left[DM'_{rs}\left(\overset{*}{\Re}_{rs}, \overset{*-}{\Re_s}\right)\right]_{n\times k}$ using Eq. (3.11) in Table 4.14.

Table 4.14 the Hamming distance matrix $\Lambda'$

| Hamming distance | $HT_1$ | $HT_2$ | $HT_3$ | $HT_4$ | $HT_5$ | $HT_6$ |
|---|---|---|---|---|---|---|
| $DM'_{1s}\left(\overset{*}{\Re}_{1s}, \overset{*-}{\Re_s}\right)$ | 0.209 | 0.138 | 0.217 | 0.000 | 0.022 | 0.260 |
| $DM'_{2s}\left(\overset{*}{\Re}_{2s}, \overset{*-}{\Re_s}\right)$ | 0.364 | 0.306 | 0.466 | 0.182 | 0.605 | 0.393 |
| $DM'_{3s}\left(\overset{*}{\Re}_{3s}, \overset{*-}{\Re_s}\right)$ | 0.131 | 0.000 | 0.193 | 0.176 | 0.144 | 0.122 |
| $DM'_{4s}\left(\overset{*}{\Re}_{4s}, \overset{*-}{\Re_s}\right)$ | 0.000 | 0.120 | 0.124 | 0.337 | 0.229 | 0.000 |
| $DM'_{5s}\left(\overset{*}{\Re}_{5s}, \overset{*-}{\Re_s}\right)$ | 0.296 | 0.436 | 0.000 | 0.354 | 0.000 | 0.225 |

(4) Figure out the entropy $E_s\ (s = 1, 2, \cdots, 6)$ of each attribute $HT_s$ using Eq. (3.12).

Table 4.15 the entropy $E_s$ of each attribute

| | $HT_1$ | $HT_2$ | $HT_3$ | $HT_4$ | $HT_5$ | $HT_6$ |
|---|---|---|---|---|---|---|
| $E_s$ | 0.821 | 0.778 | 0.785 | 0.839 | 0.625 | 0.814 |
| $1-E_s$ | 0.179 | 0.222 | 0.215 | 0.161 | 0.375 | 0.186 |

(5) Compute out the original relative weight $\varpi_s\left(s=1,2,\cdots,6\right)$ using Eq. (3.13).

Table 4.16 the original attribute weight $\varpi_s$

| Original attribute weight | $HT_1$ | $HT_2$ | $HT_3$ | $HT_4$ | $HT_5$ | $HT_6$ |
|---|---|---|---|---|---|---|
| $\varpi_s$ | 0.134 | 0.166 | 0.160 | 0.121 | 0.280 | 0.139 |

**Step 4.** Determine the average solution $AV_s\left(s=1,2,\cdots,k\right)$ of each attribute $HT_s\left(s=1,2,\cdots k\right)$ using Eq. (3.14) in Table 4.17.

Table 4.17 the average solution $AV_s$

| Average solution | $HT_1$ | $HT_2$ | $HT_3$ |
|---|---|---|---|
| $AV_s$ | $\langle[0.607,0.720],[0.142,0.205]\rangle$ | $\langle[0.256,0.309],[0.479,0.595]\rangle$ | $\langle[0.528,0.645],[0.217,0.282]\rangle$ |
| Average solution | $HT_4$ | $HT_5$ | $HT_6$ |
| $AV_s$ | $\langle[0.281,0.332],[0.465,0.567]\rangle$ | $\langle[0.608,0.717],[0.145,0.203]\rangle$ | $\langle[0.460,0.565],[0.264,0.324]\rangle$ |

**Step 5.** Based on the IVIFN $\Re_{rs}^{*}$, $AV_s$ and Eq. (3.3), the relative attribute $g'_{rs}\left(\varpi_s\right)$ is calculated by Eq. (3.15), where parameter: $\alpha=0.61,\beta=0.69$.

Table 4.18 the relative attribute weight

| | $HT_1$ | $HT_2$ | $HT_3$ | $HT_4$ | $HT_5$ | $HT_6$ |
|---|---|---|---|---|---|---|
| $HL_1$ | 0.202 | 0.230 | 0.226 | 0.190 | 0.314 | 0.219 |
| $HL_2$ | 0.215 | 0.238 | 0.234 | 0.190 | 0.308 | 0.219 |
| $HL_3$ | 0.202 | 0.230 | 0.226 | 0.190 | 0.314 | 0.207 |
| $HL_4$ | 0.202 | 0.230 | 0.226 | 0.204 | 0.314 | 0.207 |
| $HL_5$ | 0.215 | 0.238 | 0.226 | 0.204 | 0.314 | 0.219 |

步骤 **6.** Obtain the $PDA'_{rs}$ and $NDA'_{rs}$ , where $\gamma = 0.88, \delta = 0.88, \rho = 2.25$.

Table 4.19 the positive distance $PDA'$

| | $HT_1$ | $HT_2$ | $HT_3$ | $HT_4$ | $HT_5$ | $HT_6$ |
|---|---|---|---|---|---|---|
| $HL_1$ | 0.000 | 0.000 | 0.000 | 0.000 | 0.000 | 0.387 |
| $HL_2$ | 1.436 | 0.849 | 2.131 | 0.000 | 1.412 | 1.444 |
| $HL_3$ | 0.000 | 0.000 | 0.000 | 0.000 | 0.000 | 0.000 |
| $HL_4$ | 0.000 | 0.000 | 0.000 | 0.508 | 0.000 | 0.000 |
| $HL_5$ | 0.770 | 1.840 | 0.000 | 0.847 | 0.000 | 0.120 |

Table 4.20 the negative distance $NDA'$

| | $HT_1$ | $HT_2$ | $HT_3$ | $HT_4$ | $HT_5$ | $HT_6$ |
|---|---|---|---|---|---|---|
| $HL_1$ | 0.740 | 1.824 | 0.775 | 2.566 | 2.068 | 0.000 |
| $HL_2$ | 0.000 | 0.000 | 0.000 | 0.418 | 0.000 | 0.000 |
| $HL_3$ | 2.464 | 4.251 | 1.333 | 0.491 | 1.023 | 2.170 |
| $HL_4$ | 5.126 | 2.151 | 2.799 | 0.000 | 0.384 | 4.124 |
| $HL_5$ | 0.000 | 0.000 | 5.184 | 0.000 | 2.240 | 0.000 |

**Step 7.** Calculate the weighted positive and negative distance ($SP'_r$ and $SN'_{rs}$) using Eq. (3.18) and (3.19).

**Step 8.** Obtain the normalized weighted positive and negative distance ($NSP'_r$ and $NSN'_r$) using Eq. (3.20) and (3.21).

Table 4.21 the normalized weighted distance

| the normalized weighted distance | $HL_1$ | $HL_2$ | $HL_3$ | $HL_4$ | $HL_5$ |
|---|---|---|---|---|---|
| $NSP'_r$ | 0.048 | 1.000 | 0.000 | 0.059 | 0.456 |
| $NSN'_r$ | 0.400 | 0.975 | 0.158 | 0.000 | 0.403 |

**Step 9.** Calculate the overall assessment score $S'_r$ of each alternative $HL_r$ in Table 4.22.

Table 4.22 the overall assessment score $S'$

| Overall assessment score | $HL_1$ | $HL_2$ | $HL_3$ | $HL_4$ | $HL_5$ |
|---|---|---|---|---|---|
| $S'_r$ | 0.224 | 0.987 | 0.079 | 0.029 | 0.429 |

It follows from the above: $S'_1(HL_1)=0.224$, $S'_2(HL_2)=0.987$, $S'_3(HL_3)=0.079$, $S'_4(HL_4)=0.029$, $S'_5(HL_5)=0.429$.

**Step 10.** Based on the overall assessment score $S'_r(r=1,2,\cdots,5)$, the ranking results of five alternatives are listed as follows:

$$HL_2 > HL_5 > HL_1 > HL_3 > HL_4$$

4.3 **Sensitivity analysis**

In this section, we incorporate the effects of CPT on the model and refer to the parameter value given by Tversky &Kahneman: the parameter in weight function $\alpha=0.61,\beta=0.69$ and value function $\gamma=0.88,\delta=0.88,\rho=2.25$, and get the final decision-making results. At this time, it is observed that five parameters are involved in Eq. (3.15), (3.16) and (3.17). Naturally, we consider whether the change of parameter value will affect the decision result. Therefore, in order to explore the robustness and effectiveness of IVIF-CPT-EDAS model, next we discuss the influence of individual parameter changes on our decision results.

**4.3.1 the sensitivity of parameter $\alpha$ in weight function**

When $\beta=0.69$ and the parameters in value function are invariant, the overall assessment score and alternative ranking results is shown in Table 4.23.

Table 4.23 the calculation results with different value of $\alpha$ in weight function

| $\alpha$ | $HL_1$ | $HL_2$ | $HL_3$ | $HL_4$ | $HL_5$ | Ranking results | The optimal solution |
|---|---|---|---|---|---|---|---|
| 0.61 | 0.224 | 0.987 | 0.079 | 0.029 | 0.429 | $HL_2 > HL_5 > HL_1 > HL_3 > HL_4$ | $HL_2$ |
| 0.05 | 0.228 | 0.987 | 0.079 | 0.039 | 0.458 | $HL_2 > HL_5 > HL_1 > HL_3 > HL_4$ | $HL_2$ |
| 0.15 | 0.227 | 0.987 | 0.079 | 0.037 | 0.453 | $HL_2 > HL_5 > HL_1 > HL_3 > HL_4$ | $HL_2$ |
| 0.25 | 0.227 | 0.987 | 0.079 | 0.035 | 0.448 | $HL_2 > HL_5 > HL_1 > HL_3 > HL_4$ | $HL_2$ |
| 0.35 | 0.226 | 0.987 | 0.079 | 0.034 | 0.443 | $HL_2 > HL_5 > HL_1 > HL_3 > HL_4$ | $HL_2$ |
| 0.45 | 0.225 | 0.987 | 0.079 | 0.032 | 0.438 | $HL_2 > HL_5 > HL_1 > HL_3 > HL_4$ | $HL_2$ |
| 0.55 | 0.224 | 0.987 | 0.079 | 0.030 | 0.433 | $HL_2 > HL_5 > HL_1 > HL_3 > HL_4$ | $HL_2$ |
| 0.65 | 0.224 | 0.987 | 0.079 | 0.029 | 0.427 | $HL_2 > HL_5 > HL_1 > HL_3 > HL_4$ | $HL_2$ |
| 0.75 | 0.223 | 0.987 | 0.079 | 0.027 | 0.422 | $HL_2 > HL_5 > HL_1 > HL_3 > HL_4$ | $HL_2$ |
| 0.85 | 0.222 | 0.987 | 0.079 | 0.026 | 0.417 | $HL_2 > HL_5 > HL_1 > HL_3 > HL_4$ | $HL_2$ |
| 0.95 | 0.222 | 0.987 | 0.079 | 0.025 | 0.411 | $HL_2 > HL_5 > HL_1 > HL_3 > HL_4$ | $HL_2$ |

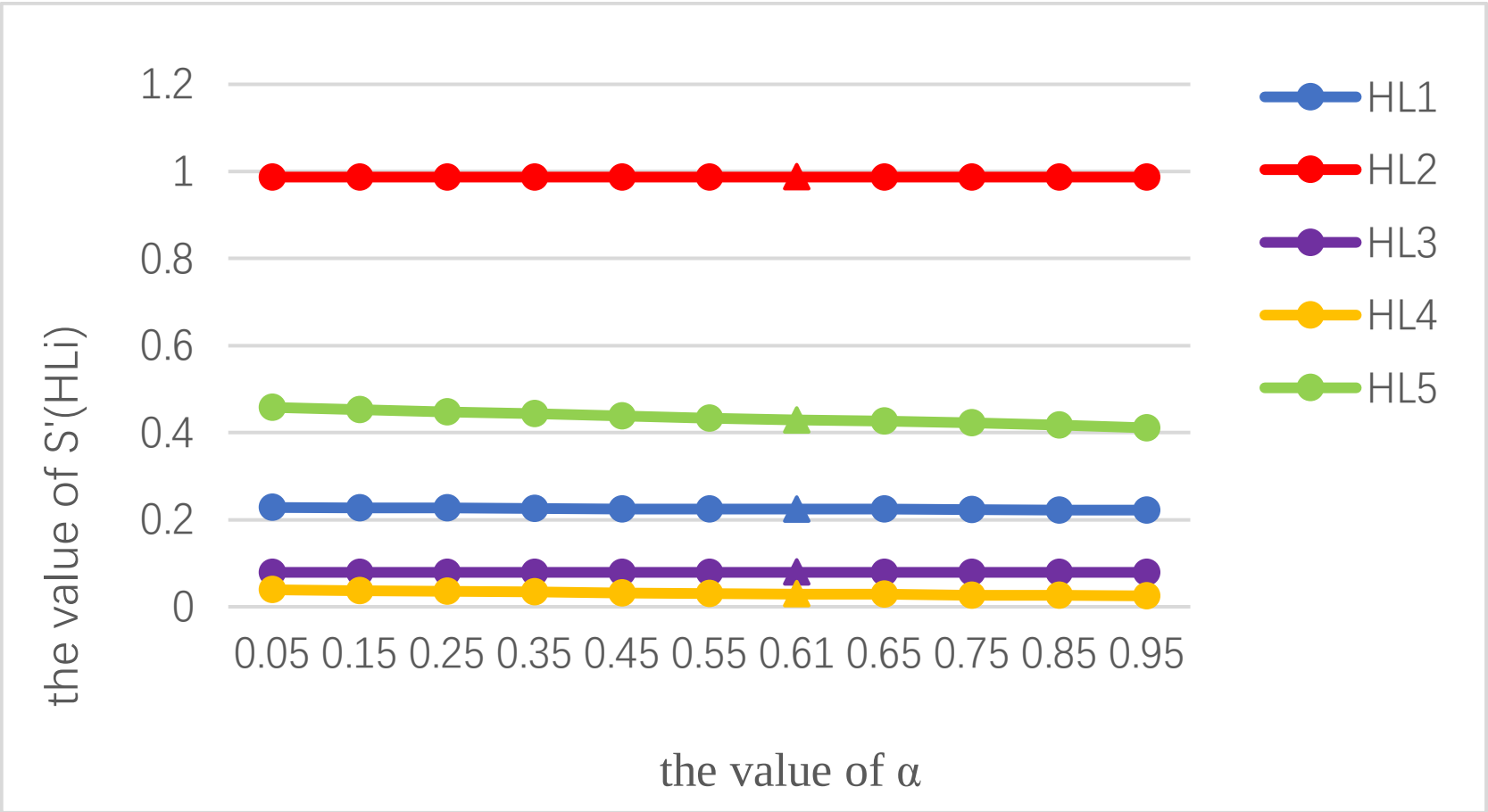

Figure 4.1 the overall assessment score with different value of parameter $\alpha$ in weight function

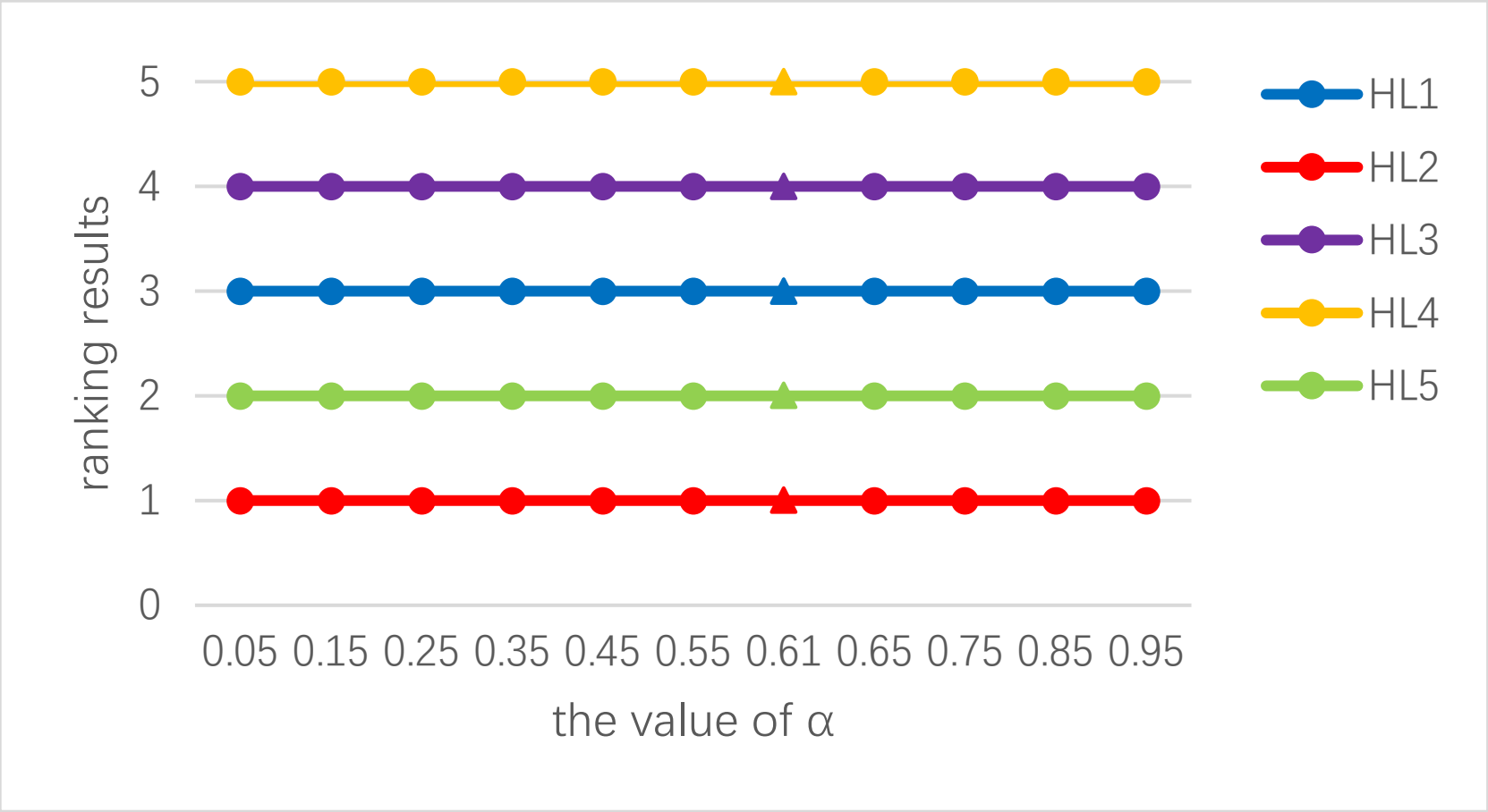

Figure 4.2 the ranking results of the different value of parameter $\alpha$ in weight function

According to Table 4.23, Figure 4.1 and Figure 4.2, different parameters have little influence on the calculation results of IVIF-CPT-EDAS model. With the value of parameter $\alpha$ increase, the overall evaluation scores of alternative schemes $HL_2$ and $HL_3$ remain unchanged, while the overall evaluation scores of other alternative schemes gradually decrease. The order of the options remains the same, the best option is $HL_2$, the worst option is $HL_4$.

4.3.2 the sensitivity analysis of parameter $\beta$ in weight function

When $\alpha = 0.61$ and the value of parameters in value function are invariant, the overall evaluation score and ranking results are shown in Table 4.24.

Table 4.24 the calculation results of different value of parameter $\beta$ in weight function

| $\beta$ | $HL_1$ | $HL_2$ | $HL_3$ | $HL_4$ | $HL_5$ | Ranking results | The optimal solution |
|---|---|---|---|---|---|---|---|
| 0.69 | 0.224 | 0.987 | 0.079 | 0.029 | 0.429 | $HL_2 > HL_5 > HL_1 > HL_3 > HL_4$ | $HL_2$ |
| 0.05 | 0.259 | 0.985 | 0.106 | 0.029 | 0.492 | $HL_2 > HL_5 > HL_1 > HL_3 > HL_4$ | $HL_2$ |
| 0.15 | 0.254 | 0.985 | 0.102 | 0.029 | 0.484 | $HL_2 > HL_5 > HL_1 > HL_3 > HL_4$ | $HL_2$ |
| 0.25 | 0.251 | 0.986 | 0.098 | 0.029 | 0.475 | $HL_2 > HL_5 > HL_1 > HL_3 > HL_4$ | $HL_2$ |
| 0.35 | 0.246 | 0.986 | 0.094 | 0.029 | 0.465 | $HL_2 > HL_5 > HL_1 > HL_3 > HL_4$ | $HL_2$ |
| 0.45 | 0.240 | 0.986 | 0.090 | 0.029 | 0.455 | $HL_2 > HL_5 > HL_1 > HL_3 > HL_4$ | $HL_2$ |
| 0.55 | 0.238 | 0.987 | 0.086 | 0.029 | 0.445 | $HL_2 > HL_5 > HL_1 > HL_3 > HL_4$ | $HL_2$ |
| 0.65 | 0.227 | 0.987 | 0.081 | 0.029 | 0.434 | $HL_2 > HL_5 > HL_1 > HL_3 > HL_4$ | $HL_2$ |
| 0.75 | 0.219 | 0.988 | 0.076 | 0.029 | 0.423 | $HL_2 > HL_5 > HL_1 > HL_3 > HL_4$ | $HL_2$ |
| 0.85 | 0.211 | 0.988 | 0.071 | 0.029 | 0.411 | $HL_2 > HL_5 > HL_1 > HL_3 > HL_4$ | $HL_2$ |
| 0.95 | 0.202 | 0.988 | 0.066 | 0.029 | 0.398 | $HL_2 > HL_5 > HL_1 > HL_3 > HL_4$ | $HL_2$ |

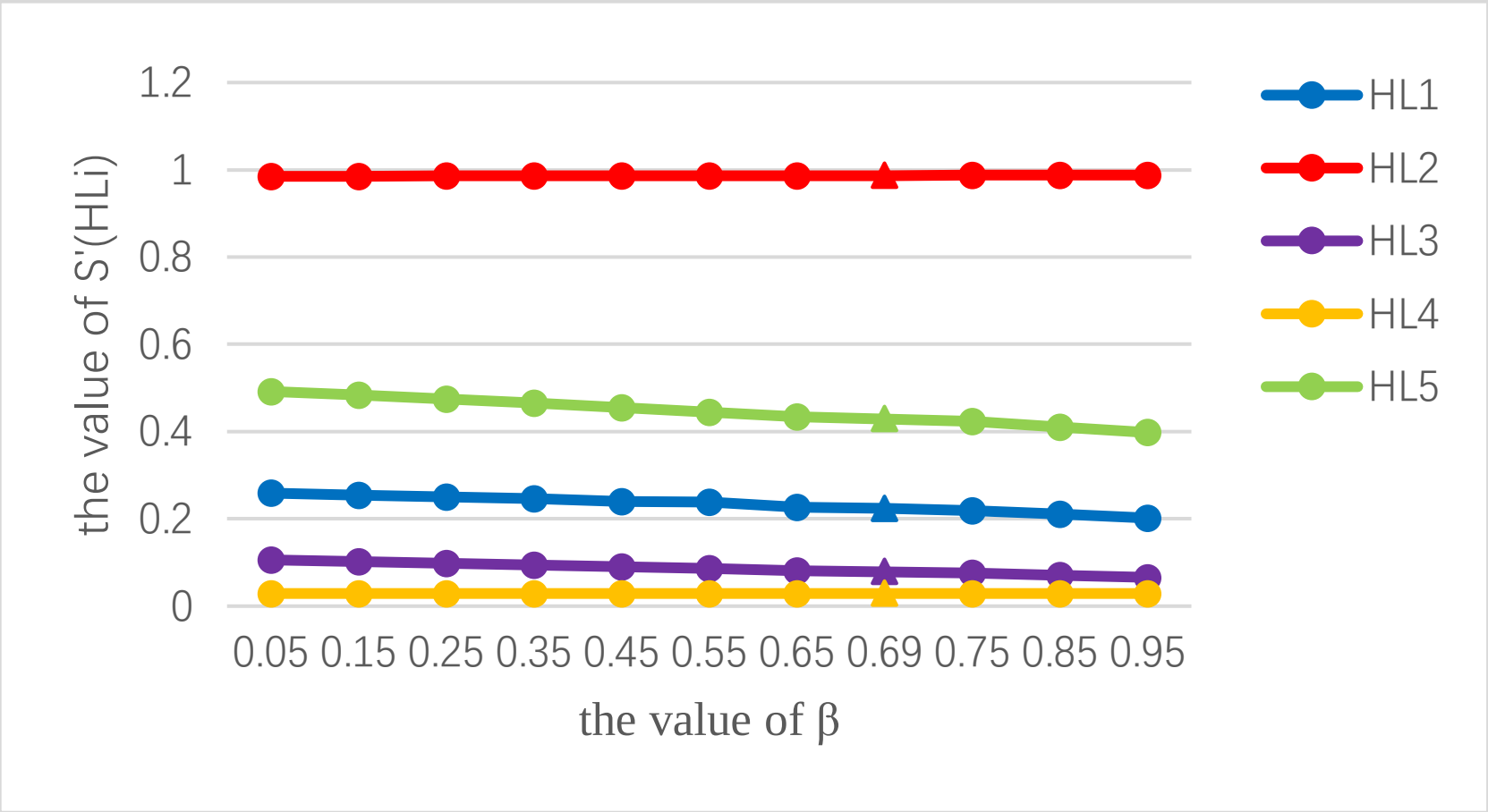

Figure 4.3 the overall assessment score with different value of parameter $\beta$ in weight function

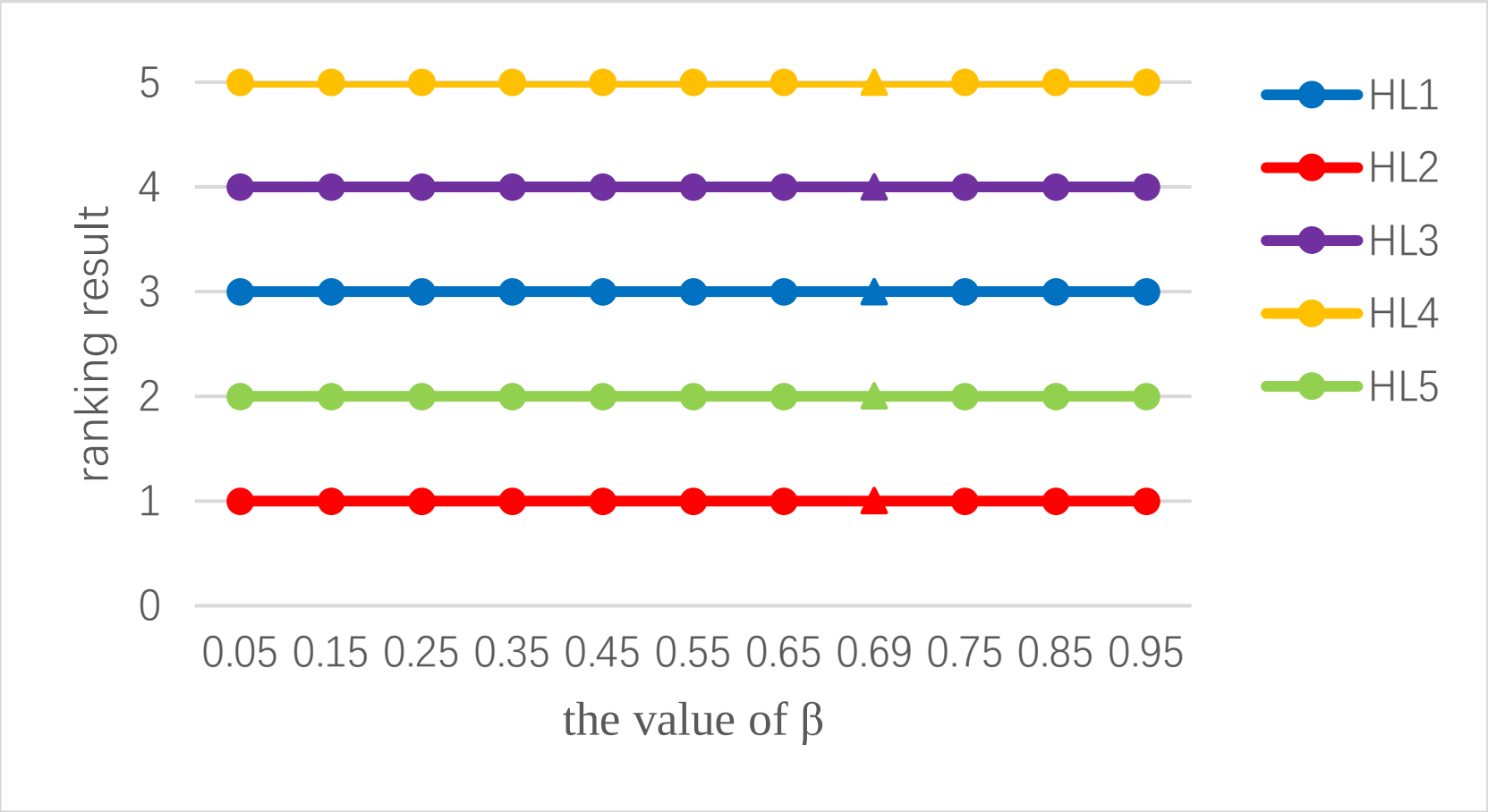

Figure 4.4 the ranking results with different value of parameter $\beta$ in weight function

According to Table 4.24, Figure 4.3 and Figure 4.4, different value of parameter $\beta$ will cause certain fluctuations to the calculation results of IVIF-CPT-EDAS model. With the value of parameter $\beta$ increase, the overall evaluation score of alternative $HL_4$ remains unchanged, the overall evaluation score of alternative $HL_2$ gradually increases, and the overall evaluation score of other alternative schemes gradually decreases, but the ranking of each scheme remains unchanged, the optimal scheme is $HL_2$, and the worst scheme is $HL_4$.

**4.3.3 the sensitivity analysis of parameter $\gamma$ in value function**

When $\delta = 0.88, \rho = 2.25$ and the parameters in weight function are in variant, the overall assessment score and ranking results are shown in Table 4.25.

Table 4.25 the calculation results with different value of parameter $\gamma$ in value function

| $\gamma$ | $HL_1$ | $HL_2$ | $HL_3$ | $HL_4$ | $HL_5$ | Ranking results | The optimal solution |
|---|---|---|---|---|---|---|---|
| 0.88 | 0.224 | 0.987 | 0.079 | 0.029 | 0.429 | $HL_2 > HL_5 > HL_1 > HL_3 > HL_4$ | $HL_2$ |
| 0.05 | 0.275 | 0.987 | 0. 079 | 0. 079 | 0.539 | $HL_2 > HL_5 > HL_1 > HL_3 = HL_4$ | $HL_2$ |
| 0.15 | 0.265 | 0.987 | 0. 079 | 0. 070 | 0.515 | $HL_2 > HL_5 > HL_1 > HL_3 > HL_4$ | $HL_2$ |
| 0.25 | 0.257 | 0.987 | 0. 079 | 0.062 | 0.495 | $HL_2 > HL_5 > HL_1 > HL_3 > HL_4$ | $HL_2$ |
| 0.35 | 0.250 | 0.987 | 0. 079 | 0.056 | 0.479 | $HL_2 > HL_5 > HL_1 > HL_3 > HL_4$ | $HL_2$ |
| 0.45 | 0.244 | 0.987 | 0. 079 | 0.049 | 0.466 | $HL_2 > HL_5 > HL_1 > HL_3 > HL_4$ | $HL_2$ |
| 0.55 | 0.238 | 0.987 | 0. 079 | 0.044 | 0.455 | $HL_2 > HL_5 > HL_1 > HL_3 > HL_4$ | $HL_2$ |
| 0.65 | 0.233 | 0.987 | 0. 079 | 0.039 | 0.446 | $HL_2 > HL_5 > HL_1 > HL_3 > HL_4$ | $HL_2$ |
| 0.75 | 0.229 | 0.987 | 0. 079 | 0.035 | 0.437 | $HL_2 > HL_5 > HL_1 > HL_3 > HL_4$ | $HL_2$ |
| 0.85 | 0.225 | 0.987 | 0. 079 | 0.031 | 0.431 | $HL_2 > HL_5 > HL_1 > HL_3 > HL_4$ | $HL_2$ |
| 0.95 | 0.222 | 0.987 | 0. 079 | 0.027 | 0.426 | $HL_2 > HL_5 > HL_1 > HL_3 > HL_4$ | $HL_2$ |

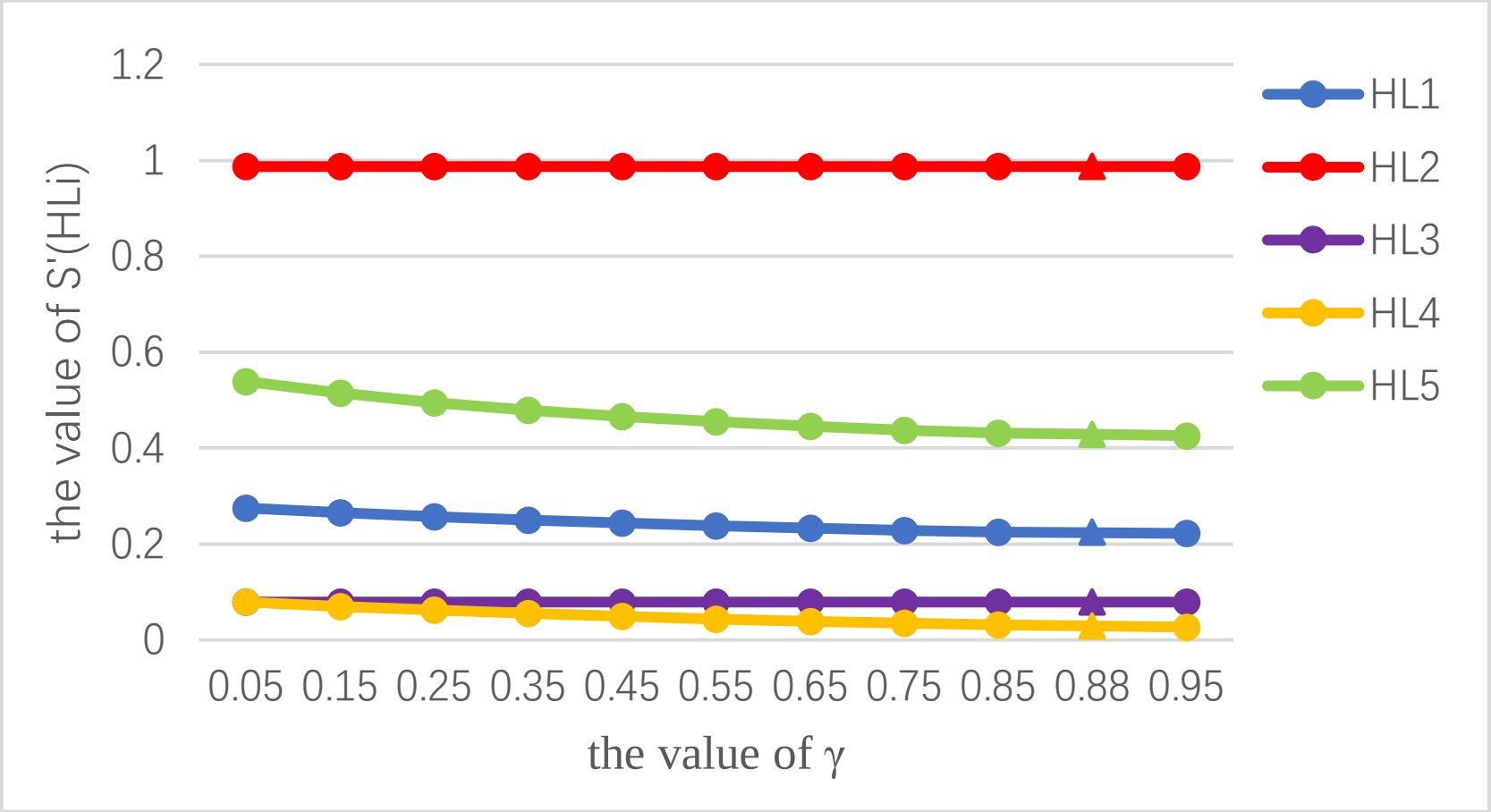

Figure 4.5 the overall assessment score with different value of parameter $\gamma$

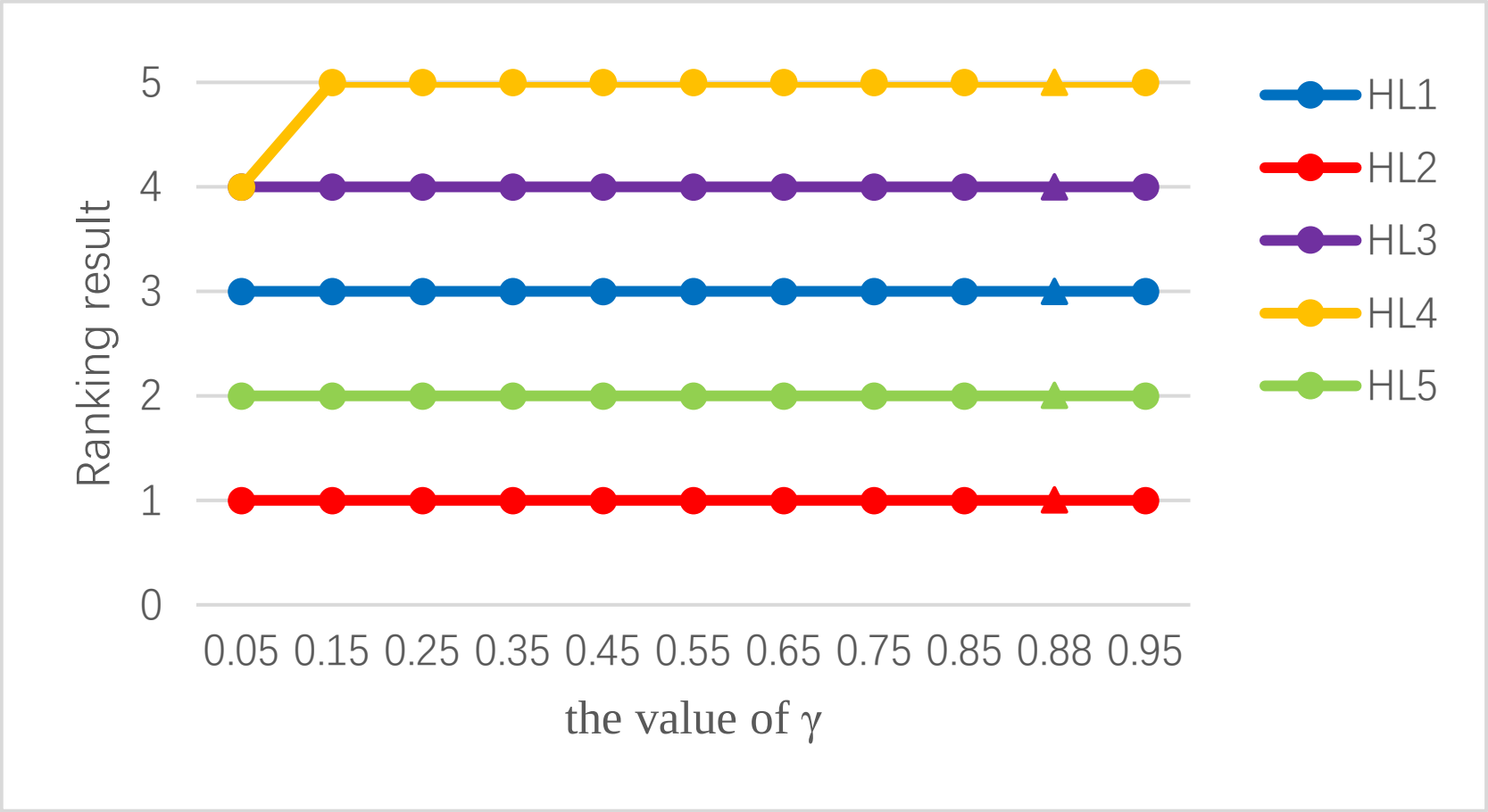

Figure 4.6 the ranking results of different value of parameter $\gamma$ in value function

According to Table 4.25, Figure 4.5 and Figure 4.6, different value of parameter $\gamma$ have little influence on the calculation results of CPT-IIF-EDAS model. With the value of parameter $\gamma$ increase, the overall evaluation scores of alternative $HL_2$ and $HL_3$ remain unchanged, while the overall evaluation scores of other alternatives gradually decrease. The order of the alternatives almost remains the same, the optimal alternative is always $HL_2$, the rest of the worst alternatives are $HL_4$. Except for $\gamma = 0.05$, the worst alternatives are $HL_3$ and $HL_4$.

**4.3.4 the sensitivity analysis with different value of parameter $\delta$ in value function**

When $\gamma = 0.88, \rho = 2.25$ and the parameters in weight function are invariant, the overall assessment score and ranking result are shown in Table 4.26.

Table 4.26 the calculation result with different value of parameter $\delta$ in value function

| $\delta$ | $HL_1$ | $HL_2$ | $HL_3$ | $HL_4$ | $HL_5$ | Ranking results | The optimal solution |
|---|---|---|---|---|---|---|---|
| 0.88 | 0.224 | 0.987 | 0.079 | 0.029 | 0.429 | $HL_2 > HL_5 > HL_1 > HL_3 > HL_4$ | $HL_2$ |
| 0.05 | 0.094 | 0.936 | 0. 000 | 0. 094 | 0.515 | $HL_2 > HL_5 > HL_1 = HL_4 > HL_3$ | $HL_2$ |
| 0.15 | 0.099 | 0.945 | 0. 000 | 0. 083 | 0.498 | $HL_2 > HL_5 > HL_1 > HL_4 > HL_3$ | $HL_2$ |
| 0.25 | 0.106 | 0.953 | 0. 000 | 0.069 | 0.482 | $HL_2 > HL_5 > HL_1 > HL_4 > HL_3$ | $HL_2$ |
| 0.35 | 0.114 | 0.960 | 0. 000 | 0.053 | 0.465 | $HL_2 > HL_5 > HL_1 > HL_4 > HL_3$ | $HL_2$ |
| 0.45 | 0.122 | 0.967 | 0. 000 | 0.034 | 0.448 | $HL_2 > HL_5 > HL_1 > HL_4 > HL_3$ | $HL_2$ |
| 0.55 | 0.144 | 0.973 | 0. 016 | 0.029 | 0.440 | $HL_2 > HL_5 > HL_1 > HL_4 > HL_3$ | $HL_2$ |
| 0.65 | 0.169 | 0.978 | 0. 036 | 0.029 | 0.436 | $HL_2 > HL_5 > HL_1 > HL_3 > HL_4$ | $HL_2$ |
| 0.75 | 0.194 | 0.983 | 0. 055 | 0.029 | 0.433 | $HL_2 > HL_5 > HL_1 > HL_3 > HL_4$ | $HL_2$ |
| 0.85 | 0.217 | 0.986 | 0. 074 | 0.029 | 0.430 | $HL_2 > HL_5 > HL_1 > HL_3 > HL_4$ | $HL_2$ |
| 0.95 | 0.240 | 0.989 | 0. 091 | 0.029 | 0.428 | $HL_2 > HL_5 > HL_1 > HL_3 > HL_4$ | $HL_2$ |

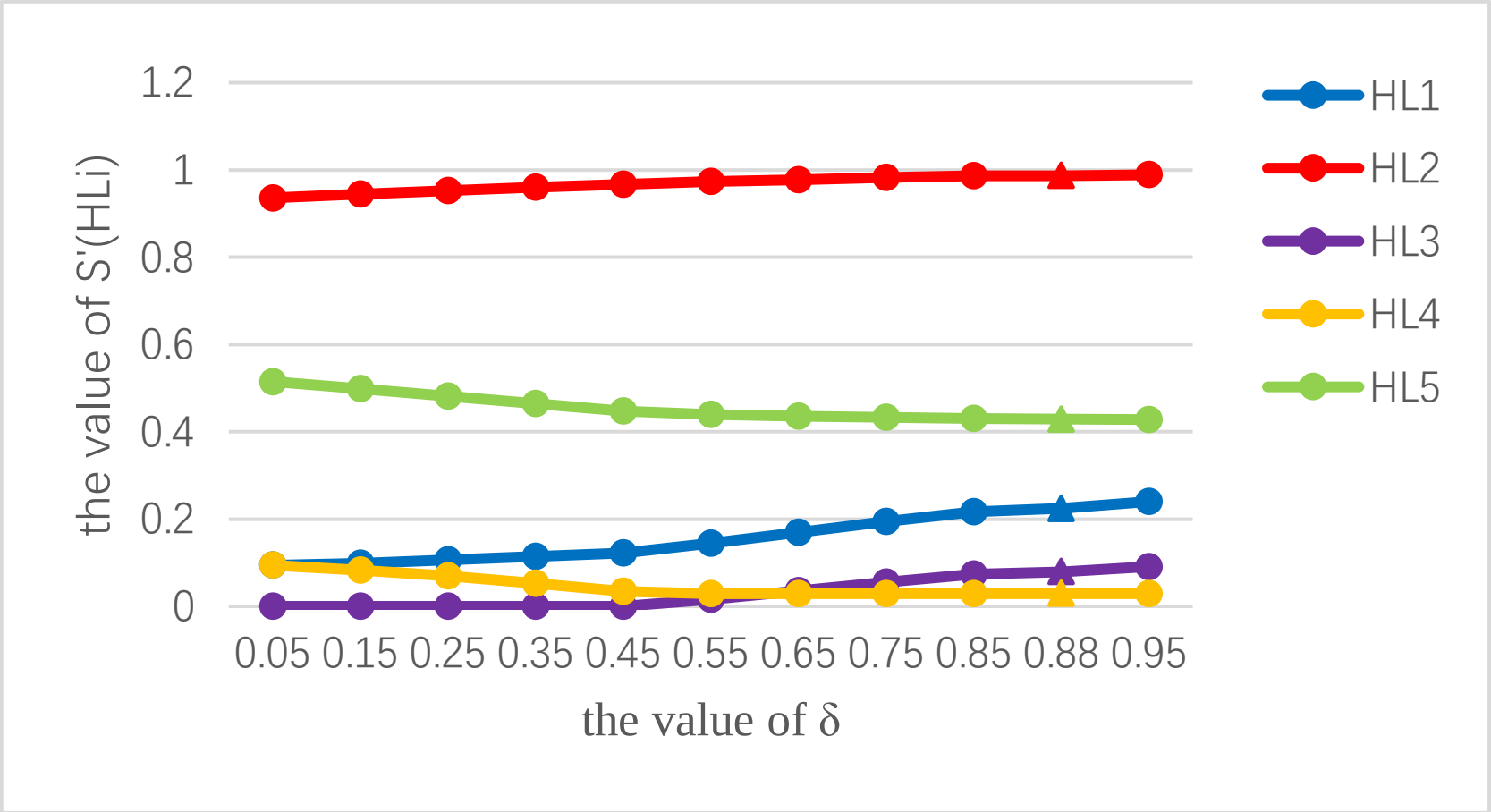

Figure 4.7 the overall assessment score with different value of parameter $\delta$ in value function

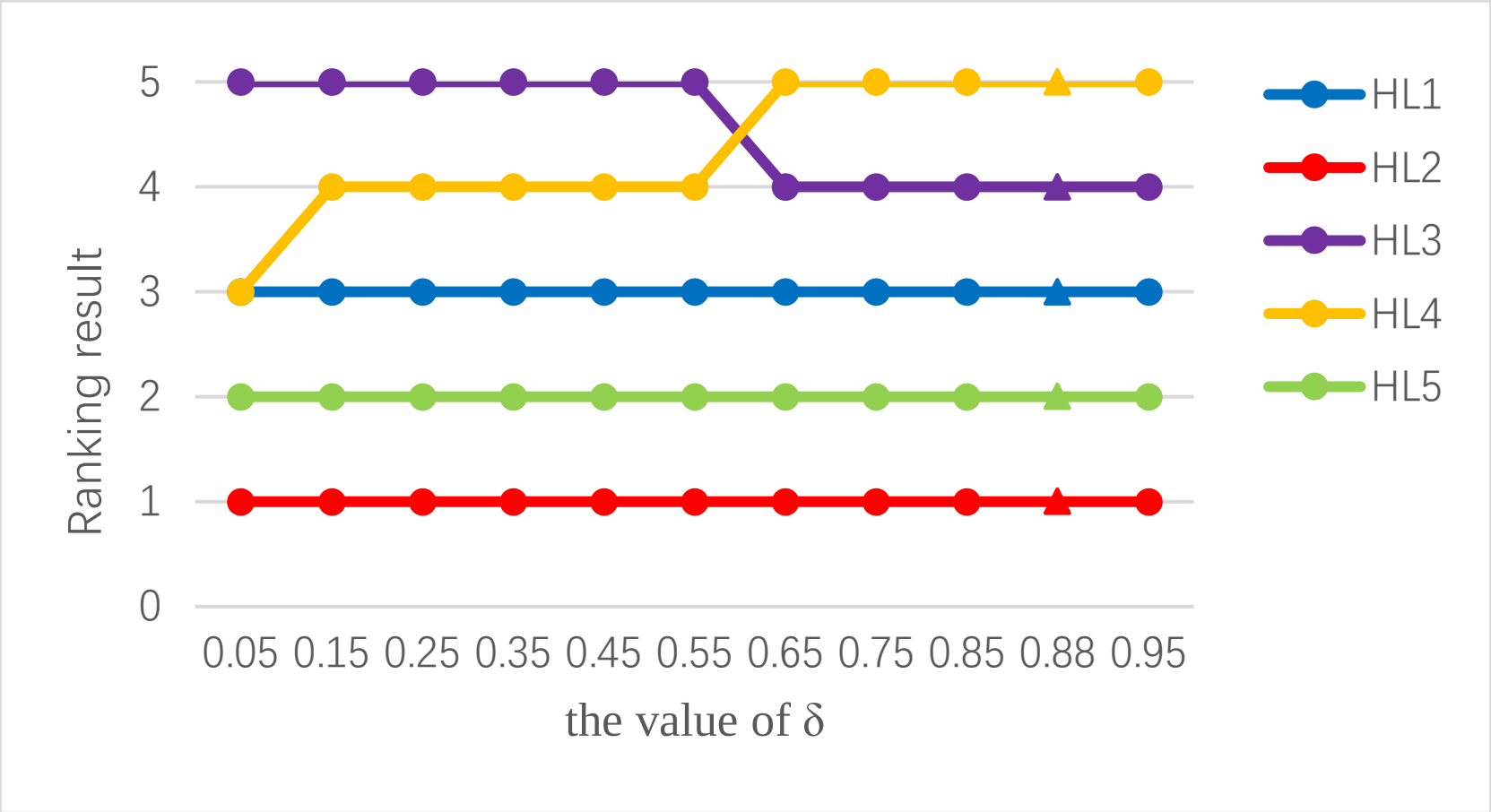

Figure 4.8 the ranking result of different value of parameter $\delta$ in value function

According to Table 4.26, Figure 4.7 and Figure 4.8, different value of parameter $\delta$ will cause certain fluctuations to the calculation results of IVIF-CPT-EDAS model. With the value of parameter $\delta$ increase, the overall evaluation score of the alternative $HL_4$ decreases gradually and then remain unchanged, and the overall evaluation score of the other alternatives decreases gradually. Although there are some changes in the ranking of alternative before and after when $\delta = 0.55$, the optimal solution is still $HL_2$ .

**4.3.5 the sensitivity analysis of parameter $\rho$ in value function**

When $\gamma = 0.88, \delta = 0.88$ and the parameters in weight function are invariant, the overall assessment score and ranking results are shown in Table 4.27.

Table 4.27 the calculation result with different value of parameter $\rho$ in value function

| $\rho$ | $HL_1$ | $HL_2$ | $HL_3$ | $HL_4$ | $HL_5$ | Ranking results | The optimal solution |
|---|---|---|---|---|---|---|---|
| 2.25 | 0.224 | 0.987 | 0.079 | 0.029 | 0.429 | $HL_2 > HL_5 > HL_1 > HL_3 > HL_4$ | $HL_2$ |
| 1.55 | 0.224 | 0.987 | 0.079 | 0.029 | 0.429 | $HL_2 > HL_5 > HL_1 > HL_3 > HL_4$ | $HL_2$ |
| 2.55 | 0.224 | 0.987 | 0.079 | 0.029 | 0.429 | $HL_2 > HL_5 > HL_1 > HL_3 > HL_4$ | $HL_2$ |
| 3.55 | 0.224 | 0.987 | 0.079 | 0.029 | 0.429 | $HL_2 > HL_5 > HL_1 > HL_3 > HL_4$ | $HL_2$ |
| 4.55 | 0.224 | 0.987 | 0.079 | 0.029 | 0.429 | $HL_2 > HL_5 > HL_1 > HL_3 > HL_4$ | $HL_2$ |
| 5.55 | 0.224 | 0.987 | 0.079 | 0.029 | 0.429 | $HL_2 > HL_5 > HL_1 > HL_3 > HL_4$ | $HL_2$ |
| 6.55 | 0.224 | 0.987 | 0.079 | 0.029 | 0.429 | $HL_2 > HL_5 > HL_1 > HL_3 > HL_4$ | $HL_2$ |
| 7.55 | 0.224 | 0.987 | 0.079 | 0.029 | 0.429 | $HL_2 > HL_5 > HL_1 > HL_3 > HL_4$ | $HL_2$ |
| 8.55 | 0.224 | 0.987 | 0.079 | 0.029 | 0.429 | $HL_2 > HL_5 > HL_1 > HL_3 > HL_4$ | $HL_2$ |
| 9.55 | 0.224 | 0.987 | 0.079 | 0.029 | 0.429 | $HL_2 > HL_5 > HL_1 > HL_3 > HL_4$ | $HL_2$ |
| 10.00 | 0.224 | 0.987 | 0.079 | 0.029 | 0.429 | $HL_2 > HL_5 > HL_1 > HL_3 > HL_4$ | $HL_2$ |

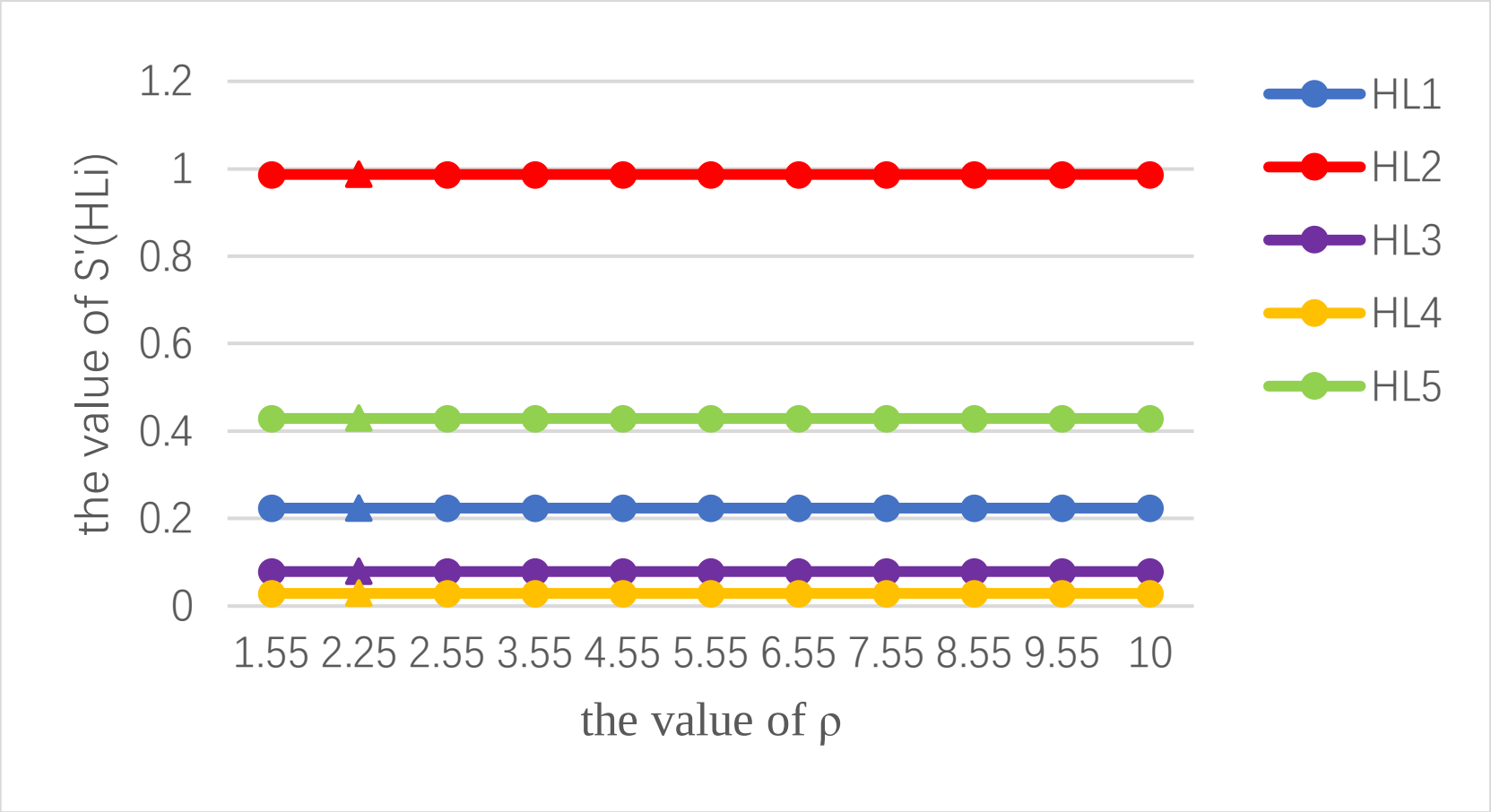

Figure 4.9 the overall assessment score with different value of parameter $\rho$ in value function

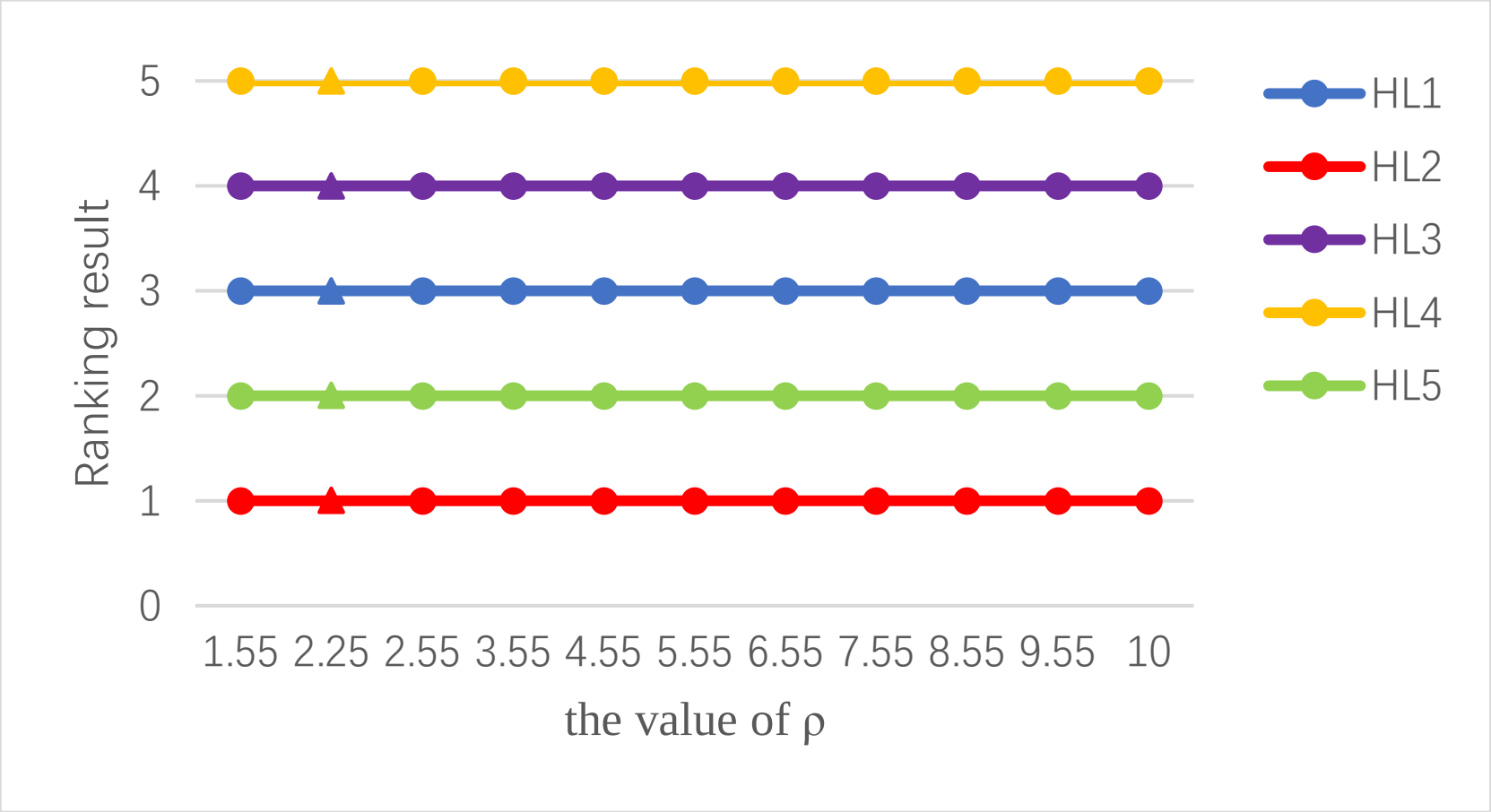

Figure 4.10 the ranking result with different value of parameter $\rho$ in value function

According to Table 4.27, Figure 4.9 and Figure 4.10, different value of parameter $\rho$ have no influence on the calculation results of IVIF-CPT-EDAS model. With the value of parameter $\rho$ increase, the overall evaluation score of all alternatives remains the same, and the ranking of each alternative remains the same. The best solution $HL_2$ remains the same, and the worst solution $HL_4$ remains the same.

According to the above observation and analysis results, although the change of parameters in the weight function and value function will cause slight fluctuations to the calculation results of the model, it will not change the optimal scheme and the suboptimal solution in the decision results, so the IVIF-CPT-EDAS model is robust and effective. At the same time, we note that the psychological preferences of DMs do not have a great impact on the decision results, which indicates that the IVIF-CPT-EDAS model has good stability.

## 5. Comparison and analysis

In this chapter, we continue to explore the effectiveness of our improved EDAS approach. Based on the same original matrices, the extended EDAS method is compared with other classical MADM methods, such as: IVIFWA operator[55], IVIF-TOPSIS method[60], IVIF-TAXONOMY method[61] and IVIF-TODIM method[62]. The corresponding calculation results are shown in Table 5.1 to 5.4.

### 5.1.1 compare with IVIFWA operator[55]

Substitute the expert weighting vector $\nu = (\nu_1, \nu_2, \cdots, \nu_5) = (0.29, 0.17, 0.19, 0.15, 0.20)$ and initial attribute weighting vector $\varpi = (\varpi_1, \varpi_2, \cdots, \varpi_6) = (0.134, 0.166, 0.160, 0.121, 0.280,0.139)$ in Tables 4.7- 4.11 into Eq. (2.16). Then the ranking result is obtained by score function and accuracy function [4] in Eq. (5.1) and (5.2). the calculation results are shown in Table 5.1.

$$SF(IVIFS) = \frac{LM - LN + RM - RN}{2} \tag{5.1}$$

$$AF(IVIFS) = \frac{LM + LN + RM + RN}{2} \tag{5.2}$$

Table 5.1 the calculation results of IVIFWA operator

| alternative | IVIFWA operator | $SF(HL_r)$ | $AF(HL_r)$ | Ranking results |
|---|---|---|---|---|
| $HL_1$ | $\langle [0.444, 0.541], [0.284, 0.349] \rangle$ | 0.176 | 0.809 | 3 |
| $HL_2$ | $\langle[0.668, 0.782],[0.103, 0.172]\rangle$ | 0.588 | 0.863 | 1 |
| $HL_3$ | $\langle[0.421, 0.517],[0.311, 0.375]\rangle$ | 0.126 | 0.812 | 4 |
| $HL_4$ | $\langle[0.403, 0.499],[0.317, 0.387]\rangle$ | 0.099 | 0.803 | 5 |
| $HL_5$ | $\langle[0.481, 0.575],[0.251, 0.325]\rangle$ | 0.240 | 0.816 | 2 |

IVIFWA operator is a classical method to solve MAGDM problems. Table 5.1 shows that the ranking of decision results of IVIFWA operator is $HL_2 > HL_5 > HL_1 > HL_3 > HL_4$ which is the same as that of the improved EDAS method in this paper. The optimal alternative is $HL_2$ and the worst alternative is $HL_4$, so the CPT-IIF-EDAS method proposed in this paper is effective.

**5.1.2 compare with the IVIF-TOPSIS method[60]**

The IVIF-TOPSIS method mainly calculates the distance between each alternative and the positive and negative ideal solution. The smaller the distance between the alternative and the positive ideal solution, the larger the distance between the alternative and the negative ideal solution, the better the alternative. The author of article [12] sorted each alternative by determining the relative closeness degree between it and the positive ideal solution. The higher the relative closeness degree, the better the solution. Substitute the expert weighting vector $\nu = (\nu_1, \nu_2, \cdots, \nu_5) = (0.29, 0.17, 0.19, 0.15, 0.20)$ and initial attribute weighting vector $\varpi = (\varpi_1, \varpi_2, \cdots, \varpi_6) = (0.134, 0.166, 0.160, 0.121, 0.280, 0.139)$ in Tables 4.7- 4.11 into Eq. (2.16). Then the ranking result is obtained by score function and accuracy function [4] in Eq. (5.1) and (5.2). the calculation results are shown in Table 5.2.

The specific processes are as follows:

**Step 1.** Calculate the weighted decision matrix $\Omega$ using the decision matrix $\aleph$, IVIFN and original attribute weight $\varpi$.

**Step 2.** Determine the IVIF-PIS $\overset{+}{\Re_s} = \left(\left[LM_s^+, RM_s^+\right], \left[LN_s^+, RN_s^+\right]\right)$ and IVIF-NIS $\overset{-}{\Re_s} = \left(\left[LM_s^-, RM_s^-\right], \left[LN_s^-, RN_s^-\right]\right)$ using weighted decision matrix $\Omega$, Eq. (5.3) and (5.4).

$$\Re_s^+ = \begin{cases} \left(\left[\max_r LM_{rs}, \max_r RM_{rs}\right], \left[\min_r LN_{rs}, \min_r RN_{rs}\right]\right), & \textit{attribute } HT_s \textit{ is positive} \\ \left(\left[\min_r LM_{rs}, \min_r RM_{rs}\right], \left[\max_r LN_{rs}, \max_r RN_{rs}\right]\right), & \textit{attribute } HT_s \textit{ is negative} \end{cases} \tag{5.3}$$

$$r = 1, 2, \cdots, n; s = 1, 2, \cdots, k$$

$$\Re_s^- = \begin{cases} \left(\left[\min_r LM_{rs}, \min_r RM_{rs}\right], \left[\max_r LN_{rs}, \max_r RN_{rs}\right]\right), & \textit{attribute } HT_s \textit{ is positive} \\ \left(\left[\max_r LM_{rs}, \max_r RM_{rs}\right], \left[\min_r LN_{rs}, \min_r RN_{rs}\right]\right), & \textit{attribute } HT_s \textit{ is negative} \end{cases} \tag{5.4}$$

$$\mathrm{r} = 1, 2, \cdots, n; s = 1, 2, \cdots, k$$

**Step 2.** Calculate the distance between each possible solution and IVIF-PIS (IVIF-NIS) using Eq. (5.5) and (5.6).

$$D_r^+ = \frac{1}{4n} \sum_{s=1}^{k} \varpi_s \begin{Bmatrix} \left|LM_{rs} - LM_s^+\right| + \left|RM_{rs} - RM_s^+\right| + \\ \left|LN_{rs} - LN_s^+\right| + \left|RN_{rs} - RN_s^+\right| \end{Bmatrix}, r = 1, 2, \cdots, n \tag{5.5}$$

$$D_r^+ = \frac{1}{4n} \sum_{s=1}^{k} \varpi_s \begin{Bmatrix} \left|LM_{rs} - LM_s^-\right| + \left|RM_{rs} - RM_s^-\right| + \\ \left|LN_{rs} - LN_s^-\right| + \left|RN_{rs} - RN_s^-\right| \end{Bmatrix}, r = 1, 2, \cdots, n \tag{5.6}$$

**Step 3.** Calculate the relative proximity degree $C_r\left(HL_r\right)$ between IVIF-PIS and each alternative using Eq. (5.7).

$$C_r = \frac{d_r^-}{d_r^+ + d_r^-}, \quad 0 \le C_r \le 1 \tag{5.7}$$

Table 5.2 the calculation results of IVIF-TOPSIS

| alternative | $D_r^+$ | $D_r^-$ | $C_r$ | Ranking results |
|---|---|---|---|---|
| $HL_1$ | 0.129 | 0.061 | 0.320 | 3 |
| $HL_2$ | 0.013 | 0.177 | 0.932 | 1 |
| $HL_3$ | 0.150 | 0.040 | 0.212 | 5 |
| $HL_4$ | 0.145 | 0.045 | 0.238 | 4 |
| $HL_5$ | 0.104 | 0.085 | 0.449 | 2 |

According to Table 5.2, we observe that the decision results of the IIF-TOPSIS method $HL_2 > HL_5 > HL_1 > HL_4 > HL_3$ are different from the worst alternative and the second-difference alternative in the decision results of the improved EDAS method in this paper, but the optimal alternative is both $HL_2$ . Therefore, the IVIF-CPT-EDAS method proposed in this paper is effective.

**5.1.3 compare with IVIF-Taxonomy[61]**

The IVIF-Taxonomy method mainly classifies and compares alternatives and uses mean and standard deviation to determine the development value of alternative schemes. The range of development value is $[0,1]$. And the more the development value of alternative schemes is away from 1, the closer it is to 0, the better the alternative. Substitute the expert weighting vector $v=\left(v_1,v_2,\cdots,v_5\right)=\left(0.29,0.17,0.19,0.15,0.20\right)$ and initial attribute weighting vector $\varpi=\left(\varpi_1,\varpi_2,\cdots,\varpi_6\right)=\left(0.134,0.166,0.160,0.121,0.280,0.139\right)$ in Tables 4.7- 4.11 into Eq. (2.16). Then the ranking result is obtained by score function and accuracy function [4] in Eq. (5.1) and (5.2). the calculation results are shown in Table 5.3.

**Step 1.** Obtain the hybrid distance matrix $M=[m_{pq}]_{k\times k}$ using Table 4.12 and Eq. (5.8).

$$m_{pq}=\frac{1}{4}\sum_{s=1}^{k}\varpi_s\left(\begin{array}{l}\left|\mu_{LM}\left(\overset{*}{\Re}_{ps}\right)-\mu_{LM}\left(\overset{*}{\Re}_{ps}\right)\right|+\left|\mu_{RM}\left(\overset{*}{\Re}_{ps}\right)-\mu_{RM}\left(\overset{*}{\Re}_{ps}\right)\right|\\+\left|v_{LN}\left(\overset{*}{\Re}_{ps}\right)-v_{LN}\left(\overset{*}{\Re}_{ps}\right)\right|+\left|v_{RN}\left(\overset{*}{\Re}_{ps}\right)-v_{RN}\left(\overset{*}{\Re}_{ps}\right)\right|\end{array}\right) \tag{5.8}$$

$$p,q=1,2,\cdots k$$

**Step 2.** The alternatives are homogenized by Eq. (5.9).

$$G=\overline{g}\pm 2H_g \tag{5.9}$$

where $\overline{g}=\frac{1}{n}\sum_{r=1}^{n}g_r$ , $H_g=\sqrt{\frac{1}{n}\sum_{r=1}^{n}\left(g_r-\overline{g}\right)}$ .

**Step 3.** Eq. (5.10) is used to determine the development mode of alternative schemes, and the optimal alternative is selected according to formulas (5.11) and (5.12), as shown in Table 5.3.

$$K_{ro}=\frac{1}{4}\sum_{s=1}^{k}\varpi_s\left(\begin{array}{l}\left|\mu_{LM}\left(\overset{*}{\Re}_{rs}\right)-\mu_{LM}\left(\overset{*+}{\Re}_{s}\right)\right|+\left|\mu_{RM}\left(\overset{*}{\Re}_{rs}\right)-\mu_{RM}\left(\overset{*+}{\Re}_{s}\right)\right|\\+\left|v_{LN}\left(\overset{*}{\Re}_{rs}\right)-v_{LN}\left(\overset{*+}{\Re}_{s}\right)\right|+\left|v_{RN}\left(\overset{*}{\Re}_{rs}\right)-v_{RN}\left(\overset{*+}{\Re}_{s}\right)\right|\end{array}\right) \tag{5.10}$$

where $\overset{*+}{\Re}_{s}=\left(\left[\max_r LN_s,\max_r LN_s\right].\left[\min_r LM_s,\min_r RM_s\right]\right)$ is the positive ideal point.

$$K=\overline{K}_{ro}+2S_{K_{ro}},r=1,2,\cdots,n \tag{5.11}$$

$$F_r=\frac{K_{ro}}{K},r=1,2,\cdots,n \tag{5.12}$$

where $\bar{K}_{ro}$ and $S_{K_{ro}}$ is the average value and variance of $K_{ro}$.

Table 5.3 the calculation results of IVIF-Taxonomy

| alternative | $F_r\left(HL_r\right)$ | Ranking results |
|---|---|---|
| $HL_1$ | 0.617 | 3 |
| $HL_2$ | 0.095 | 1 |
| $HL_3$ | 0.708 | 4 |
| $HL_4$ | 0.741 | 5 |
| $HL_5$ | 0.496 | 2 |

According to Table 5.3, we observed that the decision results of the IVIF-Taxonomy method were the same as the decision results of the improved EDAS method $HL_2 > HL_5 > HL_1 > HL_3 > HL_4$ in this paper. The optimal and the worst alternative is both $HL_2$ and $HL_4$. Therefore, the IVIF-CPT-EDAS method proposed in this paper is effective.

**5.1.4 compare with IVIF-TODIM method[62]**

The IVIF-TODIM method mainly refers to each other among alternatives, and uses the overall value to measure the dominance degree of each alternative compared with other alternatives. The larger the dominance degree, the better the solution. Substitute the expert weighting vector $\nu = \left(\nu_1, \nu_2, \cdots, \nu_5\right) = \left(0.29, 0.17, 0.19, 0.15, 0.20\right)$ and initial attribute weighting vector $\varpi = \left(\varpi_1, \varpi_2, \cdots, \varpi_6\right) = \left(0.134, 0.166, 0.160, 0.121, 0.280, 0.139\right)$ in Tables 4.7- 4.11 into Eq. (2.16). Then the ranking result is obtained by score function and accuracy function in Eq. (5.1) and (5.2). the calculation results are shown in Table 5.4.

**Step 1.** Obtain the normalized IVIF decision matrix $\aleph^{*\prime} = \left[\Re_{rs}^{*}\right]_{n\times k}$ using Table 4.12 and Eq. (5.13).

$$\Re_{rs}^{*} = \left(\left[LM_{rs}^{*\prime}, RM_{rs}^{*\prime}\right], \left[LN_{rs}^{*\prime}, RN_{rs}^{*\prime}\right]\right)$$

$$= \left( \left[ \frac{LM_{rs}^{*}}{\sum_{p=1}^{n}\left(\left(LM_{ps}^{*}\right)^{2}+\left(RM_{ps}^{*}\right)^{2}\right)^{\frac{1}{2}}}, \frac{RM_{rs}^{*}}{\sum_{r=1}^{n}\left(\left(LM_{ps}^{*}\right)^{2}+\left(RM_{ps}^{*}\right)^{2}\right)^{\frac{1}{2}}} \right], \left[ \frac{LN_{rs}^{*}}{\sum_{p=1}^{n}\left(\left(LN_{ps}^{*}\right)^{2}+\left(RN_{ps}^{*}\right)^{2}\right)^{\frac{1}{2}}}, \frac{RN_{rs}^{*}}{\sum_{r=1}^{n}\left(\left(LN_{ps}^{*}\right)^{2}+\left(RN_{ps}^{*}\right)^{2}\right)^{\frac{1}{2}}} \right] \right) \tag{5.13}$$

**Step 2.** Obtain the normalized attribute weight $\varpi' = \left(\varpi_1', \varpi_2', \cdots, \varpi_s'\right)$ using Eq. (5.14).

$$\varpi_{qs}' = \frac{\varpi_s}{\varpi_q} \tag{5.14}$$

where $\varpi_q = \max_{s}\left\{\varpi_1, \varpi_2, \cdots, \varpi_s\right\}$.

**Step 3.** The dominance degree between each alternative $HL_p(p = 1, 2, \cdots, n)$ and each alternative $HL_r(r = 1, 2, \cdots, n)$ can be calculated by Eq. (5.15), (5.16) and Eq. (5.17), where the parameter $\theta = 1$.

$$\delta_{pr}(HL_p, HL_r) = \sum_{s=1}^{k} \phi_s\left(HL_p, HL_r\right) \quad p, r = 1, 2, \cdots, n \tag{5.15}$$

$$\phi_s\left(HL_p, HL_r\right) = \begin{cases} \sqrt{\dfrac{\varpi_{qs}}{\sum_{s=1}^{k}\varpi_{qs}}} \cdot d_{prs}\left(\Re_{ps}^{*}, \Re_{rs}^{*}\right) & , \Re_{ps}^{*} > \Re_{rs}^{*} \\ 0 & , \Re_{ps}^{*} = \Re_{rs}^{*} \\ -\dfrac{1}{\theta}\sqrt{\dfrac{\sum_{s=1}^{k}\varpi_{qs}}{\varpi_{qs}}} \cdot d_{prs}\left(\Re_{ps}^{*}, \Re_{rs}^{*}\right) & , \Re_{ps}^{*} < \Re_{rs}^{*} \end{cases} \tag{5.16}$$

$$d_{prs} = \frac{1}{4}\left[\left|LM_{ps}^{*\prime} - LM_{rs}^{*\prime}\right| + \left|RM_{ps}^{*\prime} - RM_{rs}^{*\prime}\right| + \left|LN_{ps}^{*\prime} - LN_{rs}^{*\prime}\right| + \left|RN_{ps}^{*\prime} - RN_{rs}^{*\prime}\right|\right]^{\frac{1}{2}} \tag{5.17}$$

**Step 4.** Calculate the overall dominance degree $\xi_r(r = 1, 2, \cdots, n)$ of each alternative $HL_r$ using Eq. (5.18). the larger the value is, the better the alternative is.

$$\xi_r = \frac{\sum_{p=1}^{n} \delta_{pr}(HL_p, HL_r) - \min_p \delta_{pr}(HL_p, HL_r)}{\max_p \delta_{pr}(HL_p, HL_r) - \min_p \delta_{pr}(HL_p, HL_r)} \tag{5.18}$$

Table 5.4 the calculation results of IVIF-TODIM

| Alternative | $\xi_r(HL_r)$ | Ranking results |
| --- | --- | --- |
| $HL_1$ | 0.180 | 3 |
| $HL_2$ | 1.000 | 1 |
| $HL_3$ | 0.018 | 4 |
| $HL_4$ | 0.000 | 5 |
| $HL_5$ | 0.623 | 2 |

According to Table 5.4, we observe that the decision results $HL_2 > HL_5 > HL_1 > HL_3 > HL_4$ of the IVIF-TODIM method are the same as those of the improved EDAS method in this paper. The optimal alternative is both $HL_2$ and the worst alternative is both $HL_4$, so the IVIF-CPT-EDAS method proposed in this paper is effective.

5.2 Comprehensive analysis

Through the above analysis, based on the initial matrix (Table 4.7 to 4.11) in Section 4.2 and the weight of experts, all the decision-making methods in Section 5.1 were used for calculation, and the corresponding decision results were analyzed that the improved method in this paper is effective. For more intuitive comparison results, see Table 5.5, Figure 5.1 and Figure 5.2.

Table 5.5 the ranking results of different decision-making methods

| Method | order | The best choice | The worst choice |
| --- | --- | --- | --- |
| IVIFWA[55] | $HL_2 > HL_5 > HL_1 > HL_3 > HL_4$ | $HL_2$ | $HL_4$ |
| IVIF-TOPSIS[60] | $HL_2 > HL_5 > HL_1 > HL_4 > HL_3$ | $HL_2$ | $HL_4$ |
| IVIF-TOXONOMY[61] | $HL_2 > HL_5 > HL_1 > HL_3 > HL_4$ | $HL_2$ | $HL_4$ |
| IVIF-TODIM[62] | $HL_2 > HL_5 > HL_1 > HL_3 > HL_4$ | $HL_2$ | $HL_4$ |
| CPT-IVIF-EDAS | $HL_2 > HL_5 > HL_1 > HL_3 > HL_4$ | $HL_2$ | $HL_4$ |

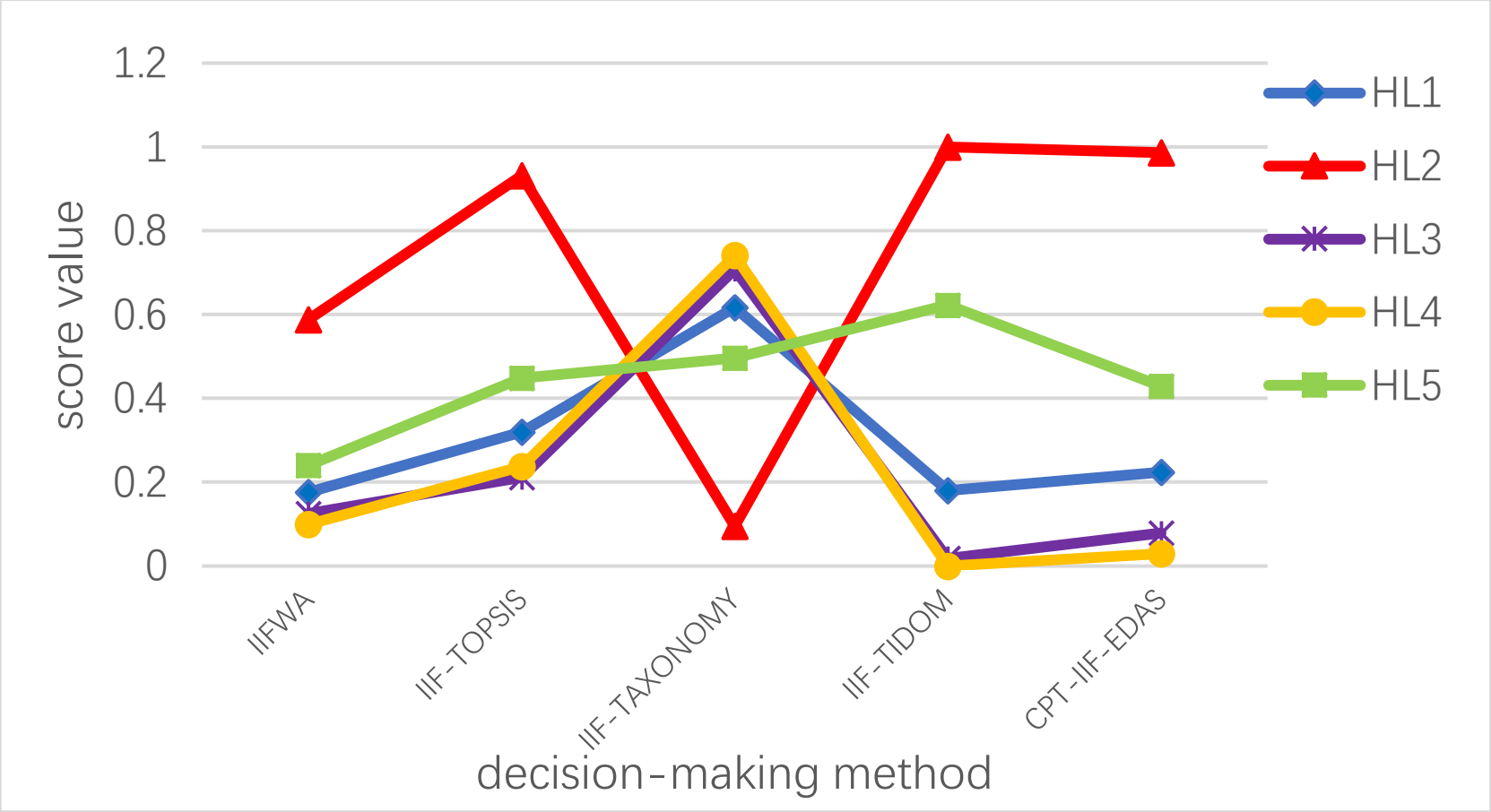

Figure 5.1 the score value of different decision-making methods

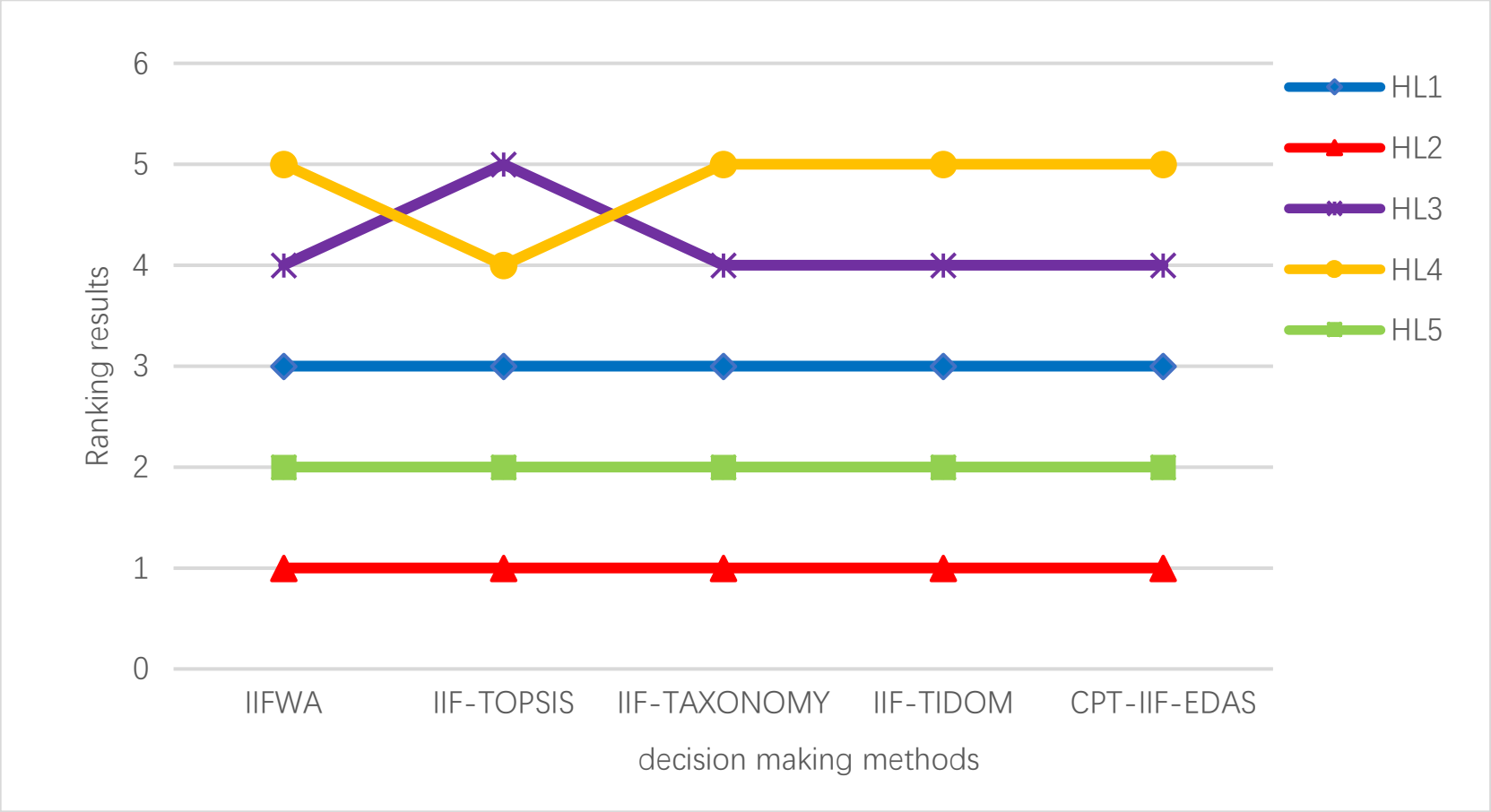

Figure 5.2 the ranking results of different decision-making methods

## 6. Conclusion

The IVIF-CPT-EDAS method studied in this paper is an extension of the MAGDM method. First, we introduce the basic definition of IVIFS, the comparison formula, the related aggregation operator, and the mixing distance of Hamming and Hausdorff. We know that EDAS method is very effective in solving decision problems with contradictory attributes. This method comprehensively determines the overall evaluation value of each alternative by calculating the positive and negative distance between each alternative and the average value under all attributes. The larger the overall evaluation value, the better the solution will be. Considering that the psychological preference of DMs has a significant influence on the new method, we then introduce the addition of CPT to the classical EDAS method, namely CPT-EDAS method. Due to the uncertainty and fuzzy characteristics of information in the actual environment, in order to retain more information and make our method more effective and scientific, we adopt in IVIFN to represent the evaluation value of each alternative attribute, that is, CPT-EDAS method is placed in IVIF-MAGDM environment. Finally, the IVIF-CPT-EDAS method is constructed. Next, based on the decision matrix of alternatives given by experts, we use the improved entropy weight method to determine the initial attribute weight, and verify the effectiveness of IVIF-CPT-EDAS method through an example of green technology venture capital. We note that a large number of parameters are designed in IVIF-CPT-EDAS method, so we conduct sensitivity analysis through parameter changes to verify the robustness of the proposed construction method. Finally, by comparing with IVIFWA operator, IVIF-TOPSIS method, IVIF-Taxonomy method and IVIF-TODIM method, the IVIF-CPT-EDAS method created in this paper has stability, accuracy and effectiveness in solving uncertain decision problems.

**Compliance with ethical standards**

**Ethical approval**

This article does not contain any studies with human participants or animals performed by any of the authors.

**Conflict of interest**

The authors declare that they have no conflict of interest.

**Data Availability**

The data used to support the findings of this study are included within the article.